\documentclass[10pt, journal]{IEEEtran}
 \usepackage[utf8]{inputenc}
\usepackage[exponent-product=.]{siunitx}

\usepackage[hyphens]{url}
\usepackage[breaklinks]{hyperref}
\hypersetup{hidelinks}
\usepackage{amsmath, 
	amsfonts, 
	amssymb, 
	bm,
	mathtools,
	amsthm,
	siunitx}
\usepackage{graphicx}
\usepackage{
		xcolor
	}
\usepackage{booktabs}
\usepackage{multirow}
\usepackage{tikz,
	pgfplots,
	pgfplotstable,
	eso-pic}
\usetikzlibrary{patterns,
	plotmarks}
\pgfplotsset{compat=1.12}

\usepackage[english,
	onelanguage,
	ruled,
	vlined,
	longend,
	titlenotnumbered]{algorithm2e}
	
\usepackage[nospace]{cite}

\usepackage{blindtext,
	ifpdf}
\usepackage{setspace}
\usepackage{paralist}
\usepackage{filecontents}

\newcommand{\bfa}{\mathbf{a}}

\newcommand{\bfA}{\mathbf{A}}
\newcommand{\bfb}{\mathbf{b}}
\newcommand{\hbfb}{\hat{\mathbf{b}}}

\newcommand{\bfB}{\mathbf{B}}

\newcommand{\bfC}{\mathbf{C}}

\newcommand{\bfe}{\mathbf{e}}

\newcommand{\bfF}{\mathbf{F}}
\newcommand{\bfG}{\mathbf{G}}

\newcommand{\bfg}{\mathbf{g}}
\newcommand{\bfh}{\mathbf{h}}
\newcommand{\bfH}{\mathbf{H}}
\newcommand{\bfI}{\mathbf{I}}

\newcommand{\bfK}{\mathbf{K}}

\newcommand{\bfN}{\mathbf{N}}
\newcommand{\calN}{\mathcal{N}}

\newcommand{\bfn}{\mathbf{n}}

\newcommand{\bfp}{\mathbf{p}}
\newcommand{\hbfp}{\hat{\mathbf{p}}}

\newcommand{\bfP}{\mathbf{P}}

\newcommand{\bfQ}{\mathbf{Q}}

\newcommand{\bfR}{\mathbf{R}}

\newcommand{\hbfR}{\hat{\bfR}}
\newcommand{\bbR}{\mathbb{R}}

\newcommand{\bfS}{\mathbf{S}}

\newcommand{\bfu}{\mathbf{u}}

\newcommand{\bfv}{\mathbf{v}}

\newcommand{\bfw}{\mathbf{w}}

\newcommand{\bfx}{\mathbf{x}}

\newcommand{\hbfx}{\hat{\mathbf{x}}}

\newcommand{\bfy}{\mathbf{y}}

\newcommand{\hbfy}{\hat{\mathbf{y}}}
\newcommand{\bfz}{\mathbf{z}}

\newcommand{\hbfz}{\hat{\bfz}}

\newcommand{\bochi}{\protect\raisebox{1pt}{$\boldsymbol{\chi}$}}
\newcommand{\hbochi}{\hat{\bochi}}

\newcommand{\boomega}{\boldsymbol{\omega}}

\newcommand{\boxi}{\boldsymbol{\xi}}

\newcommand{\bfzero}{\mathbf{0}}

\DeclareMathOperator{\diag}{diag}

\DeclareMathOperator{\cov}{cov}

\newtheoremstyle{mystyle}
{}
{}
{\itshape}
{}
{\bfseries}
{.}
{ }
{}

\usetikzlibrary{shapes,arrows,shadows}
\pgfdeclarelayer{background}
\pgfdeclarelayer{foreground}
\pgfsetlayers{background,main,foreground}

\tikzstyle{wa} = [draw, text width=10em, fill=red!20, 
minimum height=4em, rounded corners, drop shadow, text centered]
\tikzstyle{de} = [draw, text width=10em, fill=blue!20, 
minimum height=4em, rounded corners, text centered,drop shadow]

\usepgfplotslibrary{external} 
\newtheorem*{problem}{IMU Dead-Reckoning Problem}
\usepackage{scalerel}
\usetikzlibrary{svg.path}
\definecolor{orcidlogocol}{HTML}{A6CE39}
\tikzset{
	orcidlogo/.pic={
		\fill[orcidlogocol] svg{M256,128c0,70.7-57.3,128-128,128C57.3,256,0,198.7,0,128C0,57.3,57.3,0,128,0C198.7,0,256,57.3,256,128z};
		\fill[white] svg{M86.3,186.2H70.9V79.1h15.4v48.4V186.2z}
		svg{M108.9,79.1h41.6c39.6,0,57,28.3,57,53.6c0,27.5-21.5,53.6-56.8,53.6h-41.8V79.1z M124.3,172.4h24.5c34.9,0,42.9-26.5,42.9-39.7c0-21.5-13.7-39.7-43.7-39.7h-23.7V172.4z}
		svg{M88.7,56.8c0,5.5-4.5,10.1-10.1,10.1c-5.6,0-10.1-4.6-10.1-10.1c0-5.6,4.5-10.1,10.1-10.1C84.2,46.7,88.7,51.3,88.7,56.8z};
	}
}

\newcommand\orcidicon[1]{\href{https://orcid.org/#1}{
		\includegraphics[scale=0.03]{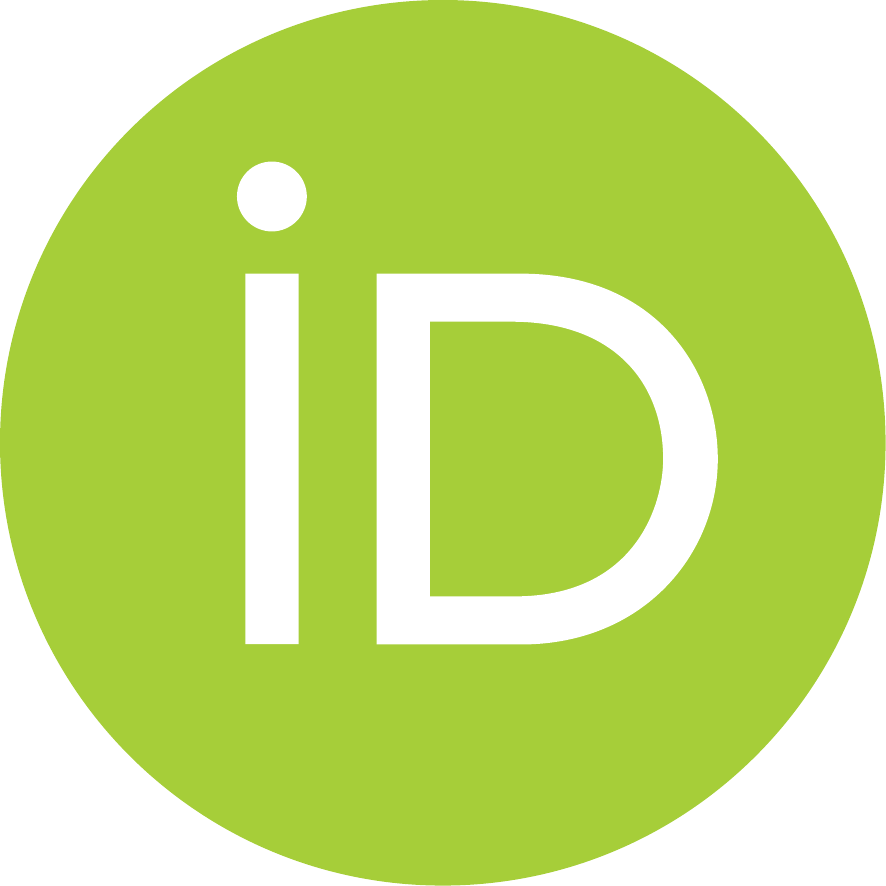}
		}}
\begin{document}
	
	\bstctlcite{IEEEexample:BSTcontrol}
	
	\title{AI-IMU Dead-Reckoning}
	
	\author{\IEEEauthorblockN{\hspace{2cm}
			Martin \textsc{Brossard}\IEEEauthorrefmark{1}\textsuperscript{\orcidicon{0000-0002-8320-6121}}
			, Axel  \textsc{Barrau}\IEEEauthorrefmark{2}\textsuperscript{\orcidicon{0000-0002-8957-8292}}
			and Silv\`ere \textsc{Bonnabel}\IEEEauthorrefmark{1}\textsuperscript{\orcidicon{0000-0002-6001-7766}}
		}
		\IEEEauthorblockA{\newline\IEEEauthorrefmark{1}MINES ParisTech, PSL Research University, Centre for Robotics, 60 Bd Saint-Michel, 75006 Paris, France}
		\IEEEauthorblockA{\IEEEauthorrefmark{2}Safran Tech, Groupe Safran, Rue des Jeunes Bois-Ch\^ateaufort, 78772, Magny Les Hameaux Cedex, France}	
	}
	
	\maketitle	
	
	\begin{abstract}
		In this paper we propose a novel accurate method for dead-reckoning of wheeled vehicles based only on an Inertial Measurement Unit  (IMU). In the context of intelligent vehicles, robust and accurate dead-reckoning based on the IMU may prove useful to correlate feeds from imaging sensors, to safely navigate through obstructions, or for safe emergency stops in the extreme case of exteroceptive sensors failure. The key components of the method are the Kalman filter and the use of deep neural networks to dynamically adapt the noise parameters of the filter. The method is tested on the KITTI odometry dataset, and our dead-reckoning inertial method based \emph{only} on the IMU accurately estimates 3D position, velocity, orientation of the vehicle and  self-calibrates the IMU biases. We achieve on average a 1.10\% translational error and the  algorithm  competes with top-ranked methods which, by contrast,  use LiDAR or  stereo vision. We make our implementation open-source at:
		\begin{center}\texttt{\url{https://github.com/mbrossar/ai-imu-dr}}\end{center}
	\end{abstract}
	
	
	\section{Introduction}\label{sec:int}
	
\begin{figure}
		\centering
		\begin{tikzpicture}[]
		\begin{axis}[height=10cm,
		width=8.5cm,
		ylabel=$y$ (\SI{}{m}),
		xlabel=$x$ (\SI{}{m}),
		xmax=800,
		xmin=-120,
		ymax=480,
		ymin=-550,
		ylabel style={xshift=-0.0cm,yshift=-0.2cm},
		xlabel style={xshift=0cm,yshift=0cm},
		legend columns=3, legend pos= north east,  
		ticks=both,
		legend entries={benchmark, IMU, \textbf{proposed}}, 
		]
		\addplot[draw = black, 
		very thick] table[x=x,y=y] {08gt.txt};
		\addplot[draw = gray, 
		very thick, forget plot] table[x=x,y=y] {08gt2.txt};
		\addplot[draw = cyan, densely dotted,very thick] table[x=x,y=y] {08imu.txt};
		\addplot[draw = green, dashdotted, very thick] table[x=x,y=y] {08nous.txt};
		\path [draw, thick, ->] (-50, -40)  node [below] {start} to[bend  right=-40,looseness=1.3, ->] (-5, -1);
		\path [draw, thick, ->] (100, 120)  node [above,text width=2cm] {GPS outage} to[bend  right=35,looseness=1.0, ->] (140, -120);
		\path [draw, thick, ->] (650, 40)  node [above] {end} to[bend  right=-40,looseness=1.3, ->] (600, -40);
		\end{axis}	
		\end{tikzpicture}
		\caption{Trajectory results on seq. \texttt{08} (drive \#28, 2011/09/30) \cite{geigerVision2013} of the KITTI dataset. The proposed method (\textcolor{green}{green}) accurately follows the benchmark trajectory for the entire sequence (\SI{4.2}{km}, \SI{9}{min}), whereas the pure IMU integration (\textcolor{cyan}{cyan})  quickly diverges.	Both methods use  only IMU signals and are initialized with the benchmark pose and velocity.  We see during the GPS outage that    occurs in this  sequence, our solution keeps estimating accurately the trajectory.\label{fig:example}} 
	\end{figure}
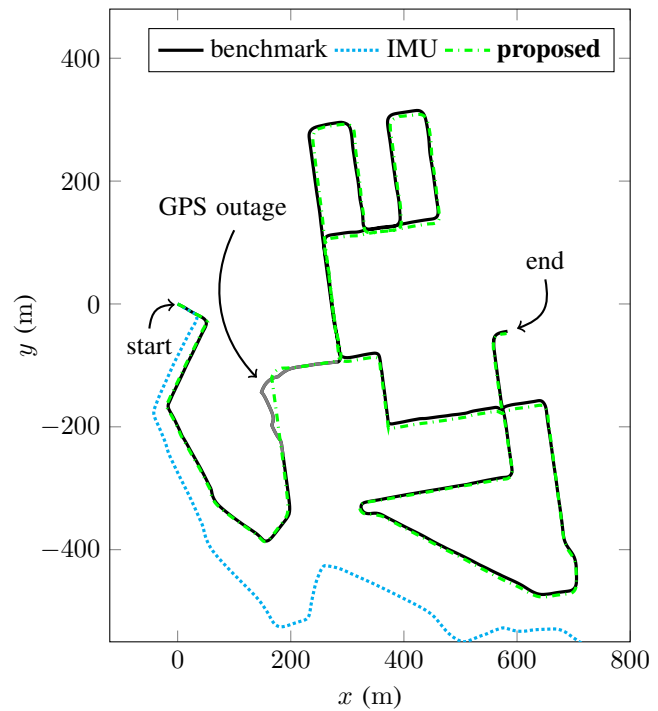

	\IEEEPARstart{I}{ntelligent} vehicles need to know where they are located in the environment, and how they are moving through it. An accurate estimate of vehicle dynamics allows validating information from imaging sensors such as lasers, ultrasonic systems, and video cameras, correlating the feeds,  and also ensuring safe motion throughout whatever may be seen along the road \cite{bressonSimultaneous2017}. Moreover, in the extreme case where an emergency stop must be performed owing to severe occlusions, lack of texture, or more generally imaging system failure, the vehicle must be able to assess accurately its dynamical motion. For all those reasons, the Inertial Measurement Unit (IMU) appears as a key component of intelligent vehicles \cite{oxtsWhy2018}.  Note that 
	Global Navigation Satellite System (GNSS) allows for global position estimation   but it suffers from phase tracking loss in densely built-up areas or through tunnels,   is sensitive to jamming, and may not be used to provide continuous accurate  and robust localization information, as exemplified by a GPS outage in the well known KITTI dataset  \cite{geigerVision2013}, see Figure \ref{fig:example}.
	
	Kalman filters are routinely used  to integrate the outputs of IMUs. When the IMU is mounted on a car, it is common practice to make the  Kalman filter   incorporate side information  about the specificity of wheeled vehicle dynamics, such as  approximately null lateral and upward velocity assumption in the form of pseudo-measurements. However,  the degree of confidence the filter should have in this side information is encoded in a covariance noise parameter which is difficult to set manually, and moreover which should dynamically adapt to the motion, e.g.,   lateral slip is larger in bends than in straight lines. Using the recent tools from the field of  Artificial Intelligence (AI), namely deep neural networks,   we propose a method to automatically learn those parameters and their dynamic adaptation   for IMU  only dead-reckoning. 
	Our contributions, and the paper's organization, are as follows:
	\begin{itemize}
		\item we introduce a state-space model for wheeled vehicles as well as simple assumptions about the motion of the car in Section \ref{sec:model};
		\item we implement a state-of-the-art Kalman filter \cite{barrauInvariant2017,barrauInvariant2018} that exploits the kinematic assumptions and combines them with the IMU outputs in a statistical way in Section \ref{sec:filter}. It yields accurate estimates of position, orientation and velocity of the car, as well as IMU  biases, along with associated uncertainty (covariance);
		\item we exploit deep learning for dynamic adaptation of covariance noise parameters of the Kalman filter  in Section \ref{sec:covnet}. This module  greatly  improves filter's robustness and accuracy, see Section \ref{sec:covnet_eval};
		\item we demonstrate the performances of the approach on the KITTI dataset \cite{geigerVision2013} in Section \ref{sec:results}. Our approach solely based on the IMU produces accurate estimates and competes with top-ranked LiDAR and stereo camera methods \cite{deschaudIMLSSLAM2018,mur-artalORBSLAM22017}; and we do not know of IMU based dead-reckoning methods capable to compete with such results;
		\item the approach is not restricted to inertial only dead-reckoning of wheeled vehicles. Thanks to the versatility of the Kalman filter, it can easily be applied for railway vehicles \cite{tschoppExperimental2019}, coupled with GNSS, the backbone for IMU self-calibration, or for using IMU as a speedometer in path-reconstruction and map-matching methods \cite{mousaInertial2018, wahlstromMapAided2016, mahmoudIntegrated2019, karlssonFuture2017}.  
	\end{itemize}

	\subsection{Relation to Previous Literature}

	Autonomous vehicle must robustly self-localize with their  embarked sensor suite which generally consists of   odometers, IMUs, radars or LiDARs, and cameras  \cite{oxtsWhy2018, karlssonFuture2017,bressonSimultaneous2017}. Simultaneous Localization And Mapping based on inertial sensors, cameras, and/or LiDARs  have enabled robust real-time localization systems, see e.g., \cite{mur-artalORBSLAM22017, deschaudIMLSSLAM2018}. Although these highly accurate solutions based on those sensors  have recently emerged, they may drift when the imaging system encounters troubles.

	As concerns wheeled vehicles, taking into account vehicle constraints and odometer measurements are known to increase the robustness of localization systems \cite{wuVINS2017, brunkerOdometry2019,zhengSE2018, buczkoSelfValidation2018}. Although quite successful, such systems  continuously process a large amount of data which is computationally demanding and energy consuming. Moreover, an autonomous vehicle should run in parallel its own robust IMU-based localization algorithm to   perform maneuvers such as  emergency stops in case of other sensors failures, or as an aid for correlation and interpretation of image feeds \cite{oxtsWhy2018}.
	
	High precision  aerial or military inertial navigation systems  achieve very small localization errors but are too costly for consumer vehicles. By contrast, low and medium-cost IMUs suffer from errors such as scale factor, axis misalignment and random walk noise, resulting in rapid localization drift \cite{kokUsing2017}. This makes the IMU-based   positioning unsuitable, even during short periods.
	
	Inertial navigation systems  have long leveraged virtual and pseudo-measurements from IMU signals, e.g. the widespread Zero velocity UPdaTe  (ZUPT) \cite{ramanandanInertial2012, dissanayakeAiding2001, brossardRINSW2019}, as  covariance adaptation   \cite{aghiliRobust2016}.  In parallel, deep learning and more generally  machine learning are gaining much interest for inertial navigation \cite{yanRIDI2018, chenIONet2018}. In \cite{yanRIDI2018} velocity is estimated using support vector regression whereas \cite{chenIONet2018} use recurrent neural networks for end-to-end inertial navigation. Those methods are promising but restricted to pedestrian dead-reckoning since they generally consider slow horizontal planar motion, and must infer velocity directly from a small sequence of IMU measurements, whereas we can afford to use larger sequences. A more general end-to-end learning approach is \cite{haarnojaBackprop2016}, which trains deep networks end-to-end in a Kalman filter. Albeit promising, the method obtains large translational error $>30 \%$ in their stereo odometry experiment. Finally, \cite{liuDeep2018} uses deep learning for estimating covariance of a local odometry algorithm that is fed into a global optimization procedure, and in \cite{brossardLearning2019} we used Gaussian processes to learn a wheel encoders error. Our conference paper \cite{brossardRINSW2019} contains preliminary ideas, albeit not concerned at all with covariance adaptation: a neural network essentially tries to detect when to perform ZUPT. 
	
	Dynamic adaptation of noise parameters in the Kalman filter is standard in the tracking literature \cite{castellaAdaptive1980}, however  adaptation rules are application dependent and are generally the result of manual ``tweaking" by engineers. Finally, in  \cite{abbeelDiscriminative2005} the authors propose to use classical machine learning techniques to to learn static noise parameters (without adaptation) of the Kalman filter, and apply it to the problem of IMU-GNSS fusion.

	\section{IMU and Problem Modelling}\label{sec:model}
	
	An inertial navigation system uses accelerometers and gyros provided by the IMU to track the orientation $\bfR_n $, velocity $\bfv_n \in \bbR^3$ and position $\bfp_n \in \bbR^3$ of a moving platform relative to a starting configuration $(\bfR_0, \bfv_0, \bfp_0)$. The orientation is encoded in a rotation matrix $\bfR_n \in SO(3)$ whose columns are the axes of  a frame attached to the vehicle. 
	
	\subsection{IMU Modelling}
	
	The IMU provides noisy and biased measurements of the instantaneous angular velocity vector $\boomega_n$ and specific acceleration $\bfa_n$ as follows \cite{kokUsing2017}
	\begin{align}
	\boomega_n^{\textsc{imu}} &= \boomega_n + \bfb_{n}^{\boomega} + \bfw_{n}^{\boomega}, \label{eq:gyro}\\
	\bfa^{\textsc{imu}}_n &= \bfa_n + \bfb_{n}^{\bfa} + \bfw_{n}^{\bfa}, \label{eq:acc}
	\end{align}
	where $\bfb_{n}^{\boomega}$, $\bfb_{n}^{\bfa}$ are quasi-constant biases and $\bfw_{n}^{\boomega}$, $\bfw_{n}^{\bfa}$ are zero-mean Gaussian noises. The biases follow a random walk
	\begin{align}
	\bfb_{n+1}^{\boomega} &= \bfb_{n}^{\boomega} + \bfw^{\bfb_{\omega}}_{n}, \label{eq:propend2}\\
	\bfb_{n+1}^{\bfa} &= \bfb_{n}^{\bfa} + \bfw^{\bfb_{\bfa}}_{n}, \label{eq:propend}
	\end{align}
	where $\bfw_{n}^{\bfb_{\boomega}}$, $\bfw^{\bfb_{\bfa}}_{n}$ are zero-mean Gaussian noises.
	
	\begin{figure}
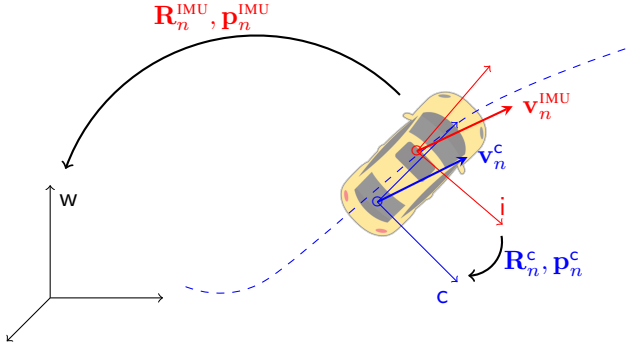

		\centering
		\include{mafigure}
		\caption{The coordinate systems that are used in the paper. The IMU pose $\left(\bfR_n^{\textsc{imu}}, \bfp_n^{\textsc{imu}}\right)$  maps vectors expressed in the IMU frame $\mathsf{i}$ (\textcolor{red}{red}) to the world frame $\mathsf{w}$ (\textcolor{black}{black}). The IMU frame is attached to the vehicle and misaligned with the car frame $\mathsf{c}$ (\textcolor{blue}{blue}). The pose between the car and inertial frames $\left(\bfR_n^{\mathsf{c}}, \bfp_n^{\mathsf{c}}\right)$ is unknown. IMU velocity $\bfv_n^{\textsc{imu}}$ and car velocity $\bfv_n^{\mathsf{c}}$ are respectively expressed in the world frame and in the car frame. \label{fig:frame}}
	\end{figure}

	The kinematic model is governed by the following equations
	\begin{align}
	\bfR_{n+1}^{\textsc{imu}} &= \bfR_n^{\textsc{imu}} \exp\left((\boomega_n dt)_\times\right), \label{eq:prop1}\\
	\bfv_{n+1}^{\textsc{imu}} &= \bfv_n^{\textsc{imu}} + \left(\bfR_n^{\textsc{imu}}\bfa_n + \bfg \right)dt, \label{eq:prop2}\\
	\bfp_{n+1}^{\textsc{imu}} &= \bfp_n^{\textsc{imu}} +  \bfv_n^{\textsc{imu}} dt,\label{eq:prop3}
	\end{align}
	between two discrete time instants sampling at $dt$, where we let  the IMU velocity be $\bfv_n^{\textsc{imu}}\in\mathbb R^3$ and its position  $\bfp_n^{\textsc{imu}}\in \mathbb R^3$ in the world frame. $\bfR_n^{\textsc{imu}}\in SO(3)$ is the $3\times3$ rotation matrix that represents the IMU orientation, i.e. that maps the IMU frame to the world frame, see Figure \ref{fig:frame}. Finally $(\mathbf{y})_\times$ denotes the skew symmetric matrix associated with cross product with $\mathbf{y}\in\mathbb R^3$.  The true angular velocity $\boomega_n \in \bbR^3$ and the true specific acceleration $\bfa_n \in \bbR^3$ are the inputs of the system \eqref{eq:prop1}-\eqref{eq:prop3}. In our application scenarios, the effects of earth rotation    and  Coriolis acceleration are ignored, Earth is considered flat, and the gravity vector $\bfg \in \bbR^3$ is a known constant.

	All sources of error displayed in \eqref{eq:gyro} and  \eqref{eq:acc}  are  harmful since a simple implementation of \eqref{eq:prop1}-\eqref{eq:prop3} leads to  a triple integration of raw  data, which is much more harmful that the unique integration of differential wheel speeds \cite{karlssonFuture2017}.  Indeed, a bias of order $\epsilon$ has an impact of order $\epsilon t^2/2$ on the position after $t$ seconds, leading to potentially huge drift.

	\subsection{Problem Modelling}
	
	We distinguish between  three different frames, see Figure \ref{fig:frame}: $i$) the static world frame, $\mathsf{w}$; $ii$) the IMU frame, $\mathsf{i}$, where \eqref{eq:gyro}-\eqref{eq:acc} are measured; and $iii$) the car frame, $\mathsf{c}$. The car frame is an ideal frame attached to the car,  that will be estimated online and plays a key role in our approach. Its orientation w.r.t. ${\mathsf{i}}$ is denoted $\bfR_n^{\mathsf{c}} \in SO(3)$  and its origin denoted $\bfp_n^{\mathsf{c}} \in \bbR^3$ is the car to IMU level arm. 
	In the rest of the paper, we tackle the following problem:
	\begin{problem}
		Given an initial known configuration $\left(\bfR_0^{\textsc{imu}}, \bfv^{\textsc{imu}}_0, \bfp_0^{\textsc{imu}}\right)$, perform in real-time IMU dead-reckoning, i.e. estimate the IMU and car variables \begin{align}\bfx_n:=\left(\bfR_n^{\textsc{imu}},~\bfv_n^{\textsc{imu}},~\bfp_n^{\textsc{imu}},~\bfb_{n}^{\boomega}, ~\bfb_{n}^{\bfa},~ \bfR_n^{\mathsf{c}}, ~\bfp_n^{\mathsf{c}}\right)\label{statedef}\end{align}using only the inertial measurements $\boomega_n^{\textsc{imu}}$ and $\bfa_n^{\textsc{imu}}$.
	\end{problem}

	\section{Kalman Filtering with Pseudo-Measurements}
	
	The Extended Kalman Filter (EKF) was first implemented in the Apollo program to localize the space capsule, and is now pervasively used in the localization industry, the radar industry, and robotics. It starts from a dynamical discrete-time non-linear law of the form
	\begin{align}
	\bfx_{n+1}=f(\bfx_n,~\bfu_n)+\bfw_n
	\label{eq:covpropRR}
	\end{align}
	where $\bfx_n$ denotes the state to be estimated, $\bfu_n$ is an input, and $\bfw_n$ is the process noise which is assumed Gaussian with zero mean and covariance matrix $\bfQ_n$.  Assume side information is in the form of loose equality constraints $h(\bfx_n)\approx \bfzero$ is available. It is then customary to  generate  a fictitious observation from the constraint function:
	\begin{align}
	\bfy_{n}=h(\bfx_{n})+\bfn_n,
	\label{eq:covpropRRR}
	\end{align}and to feed the filter with the information  that $\bfy_{n}=\bfzero$ (pseudo-measurement) as first advocated by \cite{tahkTarget1990}, see also \cite{wuVINS2017, simonKalman2010} for application to visual inertial localization and general considerations. The noise is assumed to be a centered Gaussian $\bfn_{n}\sim\mathcal N(\bfzero,\bfN_{n})$ where the covariance matrix $\bfN_{n}$ is set by the user and reflects the degree of validity of the information: the larger $\bfN_{n}$ the less confidence is put in the information.

	Starting from an initial Gaussian belief about the state, $\bfx_0\sim\mathcal N(\hat \bfx_0,\bfP_{0})$ where $\hat \bfx_0$ represents the initial estimate and the covariance matrix $\bfP_0$ the uncertainty associated to it, the EKF alternates between two steps. At the propagation step, the estimate $\hat \bfx_n$ is propagated through model  \eqref{eq:covpropRR} with noise turned off, $\bfw_n=\bfzero$, and the covariance matrix is updated through
	\begin{align}
	\bfP_{n+1} = \bfF_n \bfP_n \bfF_n^T + \bfG_n \bfQ_n \bfG_n^T, \label{eq:covprop}
	\end{align}
	where  $\bfF_n$, $\bfG_n$  are Jacobian matrices of $f(\cdot)$ with respect to $\bfx_n$ and $\bfu_n$. At the update step, pseudo-measurement is taken into account, and Kalman equations allow to update the estimate $\hat\bfx_{n+1}$ and its covariance matrix $\bfP_{n+1}$ accordingly.

	To  implement an EKF, the designer needs to determine the functions $f(\cdot)$ and $h(\cdot)$, and the associated noise matrices $\bfQ_n$ and $\bfN_n$. In this paper, noise parameters $\bfQ_n$ and $\bfN_n$ will be wholly learned by a neural network. 
	
	\subsection{Defining the Dynamical Model $f(\cdot)$}

	We now need to assess the evolution of state variables \eqref{statedef}. The evolution of   $\bfR_n^{\textsc{imu}}$, $\bfp_n^{\textsc{imu}}$, $\bfv_n^{\textsc{imu}}$, $\bfb_n^{\boomega}$ and  $\bfb_n^\bfa$ is already given by  the standard equations \eqref{eq:propend2}-\eqref{eq:prop3}.  The additional variables $\bfR_n^{\mathsf{c}}$ and $\bfp_n^{\mathsf{c}}$ represent the car frame with respect to the IMU. This car frame is rigidly attached to the car and encodes an unknown fictitious point where the pseudo-measurements of Section \ref{considered:ed} are most advantageously made. As  IMU is also rigidly attached to the car,  and $\bfR_n^{\mathsf{c}}$, $\bfp_n^{\mathsf{c}}$ represent misalignment between IMU and car frame, they are approximately constant
	\begin{align}
	\bfR_{n+1}^{\mathsf{c}} &=\bfR_n^{\mathsf{c}}\exp((\bfw_n^{\bfR^{\mathsf{c}}})_\times),  \label{eq:ass3}\\
	\bfp_{n+1}^{\mathsf{c}} &=\bfp_n^{\mathsf{c}}+  \bfw_n^{\bfp^{\mathsf{c}}}.  \label{eq:ass4}
	\end{align}
	where we let $\bfw_n^{\bfR^{\mathsf{c}}}$, $\bfw_n^{\bfp^{\mathsf{c}}}$ be centered Gaussian noises with  small covariance matrices $\sigma^{\bfR^{\mathsf{c}}} \bfI$, $\sigma^{\bfp^{\mathsf{c}}} \bfI $ that will be learned during training. Noises  $\bfw_n^{\bfR^{\mathsf{c}}}$ and  $\bfw_n^{\bfp^{\mathsf{c}}}$ encode possible small variations through time of level arm due to the lack of rigidity stemming from   dampers and shock absorbers.

	\subsection{Defining the Pseudo-Measurements $h(\cdot)$}\label{considered:ed}
	
	Consider the different frames depicted on Figure \ref{fig:frame}. The velocity of the origin point of the car frame, expressed in the car frame, writes
	\begin{align}
	\bfv^{\mathsf{c}}_n = \begin{bmatrix}
	v_n^{\mathrm{\mathrm{for}}} \\
	v_n^{\mathrm{lat}} \\
	v_n^{\mathrm{up}}
	\end{bmatrix} = \bfR^{\mathsf{c}T}_n \bfR_n^{\textsc{imu}T} \bfv_n^{\mathrm{\textsc{imu}}} + (\boomega_n)_\times \bfp_n^{\mathsf{c}},
	\end{align}
	from basic screw theory,  where $\bfp_n^{\mathsf{c}} \in \bbR^3$ is the car to IMU level arm. 
	In the car frame, we consider that the car lateral and vertical velocities are roughly null, that is, we generate two scalar pseudo observations of the form \eqref{eq:covpropRRR} that is, 
	\begin{align}
	\bfy_{n} = \begin{bmatrix}
	y_{n}^{\mathrm{lat}} \\
	y_{n}^{\mathrm{up}}
	\end{bmatrix}
	= \begin{bmatrix}
	h^{\mathrm{lat}}(\bfx_{n})+\mathrm{n}_{n}^{\mathrm{lat}} \\ 
	h^{\mathrm{up}}(\bfx_{n})+\mathrm{n}_{n}^{\mathrm{up}}
	\end{bmatrix}
	= 
	\begin{bmatrix} v_n^{\mathrm{lat}} \\
	v_n^{\mathrm{up}} 
	\end{bmatrix}+ \bfn_n,
	\label{eq:ass2}
	\end{align} 
	where the noises $\bfn_ = \left[\mathrm{n}_n^{\mathrm{lat}},\mathrm{n}_n^{\mathrm{up}}\right]^T$ are assumed centered and Gaussian with covariance matrix $\bfN_{n}\in\mathbb R^{2\times 2}$. The filter is then fed with the pseudo-measurement that $y_{n}^{\mathrm{lat}}=y_{n}^{\mathrm{up}}=0$. 
	
	Assumptions that $v_n^{\mathrm{lat}}$ and $v_n^{\mathrm{up}}$  are roughly null are common for cars moving forward on human made roads or wheeled robots moving indoor. Treating them as loose constraints, i.e., allowing the uncertainty encoded in $\bfN_{n}$ to be non strictly null, leads to much better estimates than treating them as strictly null \cite{wuVINS2017}.

	It should be duly noted the vertical velocity $v_n^{\mathrm{up}}$ is expressed in the car frame, and thus the   assumption it is roughly null generally holds for a car moving on a road even if the motion is 3D. It is quite different from assuming null vertical velocity in the world frame, which then boils down to planar horizontal motion.
	
	The main point  of the present work  is that the validity of the null lateral and vertical velocity assumptions widely vary depending on what maneuver is being performed:  for instance, $v_n^{\mathrm{lat}} $ is much larger in turns than in straight lines. The role of the noise parameter adapter of  Section \ref{sec:covnet}, based on AI techniques, will be  to dynamically assess the parameter $\bfN_{n}$ that reflects confidence in the assumptions, as a function of   past and present IMU measurements. 
	
	\subsection{The Invariant Extended Kalman Filter (IEKF)}\label{sec:filter}
	
	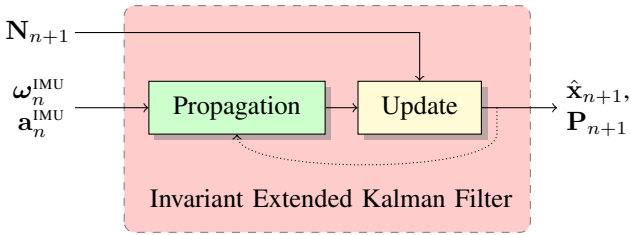
\begin{figure}[h]
		\centering
		\begin{tikzpicture}
    \node (wa) [draw, text width=6em,
    minimum height=2em, drop shadow, text centered,fill=green!20]  {Propagation};
    \path (wa.west)+(-1.5,0.0) node[text width=0.8cm,align=right] (asr2) {$\boomega_n^{\textsc{imu}}$  $\bfa_n^{\textsc{imu}}$};
    
    \path (wa.west)+(-1.5,1.0) node[text width=0.8cm,align=right] (z) {$\bfN_{n+1}$};
    
    \path  (wa.east) -- ++(1.25,0.0) node (th) [draw, text width=4em, minimum height=2em, drop shadow, text centered,fill=yellow!20]  {Update};
    \path [draw, ->] (wa.east) -- (th.west);
    \path (th.east)+(1,0) node[right,text width = 0.8cm] (vote)  {$\hbfx_{n+1}$, $\bfP_{n+1}$};
	\path [draw,densely dotted, ->] (th.east)+(0.2,0) -- ++(0.2,-0.35) to[bend  right=-90,looseness=0.4, ->] (wa.south);   

    \path [draw, ->] (asr2.east) -- (wa.west);
    \path [draw, ->] (z) -| node [above] {} (th.north);
    \path [draw, ->] (th.east) -- node [above] {} (vote.west);
        
    \path (wa.south) +(1.25,-0.85) node (asrs) {Invariant Extended Kalman Filter};
    
    \begin{pgfonlayer}{background}
        \path (wa.south)+(-1.5,-1.3) node (a) {};
        \path (wa.north)+(3.9,1) node (c) {};
        \path[rounded corners, draw=black!50, dashed, fill=red!20]
            (a) rectangle (c);           
    \end{pgfonlayer}
\end{tikzpicture}
		\caption{Structure of the IEKF. The filter uses  the noise parameter $\bfN_{n+1}$ of pseudo-measurements \eqref{eq:ass2} to yield a real time   estimate  of the state $\hbfx_{n+1}$ along with covariance $\bfP_{n+1}$. \label{fig:filter}}
	\end{figure}
	
	For inertial navigation, we advocate the use of a recent EKF variant, namely the  Invariant Extended Kalman Filter  (IEKF), see \cite{barrauInvariant2017, barrauInvariant2018},  that has recently given raise to a commercial aeronautics product \cite{barrauAligment2016} and to various successes in the field of visual inertial odometry  \cite{brossardUnscented2018, heoConsistent2018, hartleyContactAided2018}. We thus opt for an IEKF to perform the fusion between the IMU measurements \eqref{eq:gyro}-\eqref{eq:acc} and  \eqref{eq:ass2} treated as pseudo-measurements. Its architecture, which is identical to the conventional EKF's, is recapped in  Figure \ref{fig:filter}.  
	
	However, understanding in detail the IEKF \cite{barrauInvariant2017} requires  some background in Lie group geometry. The interested reader is referred to the Appendix where the exact equations are provided.

	\section{Proposed AI-IMU Dead-Reckoning}\label{sec:system}
	
	\begin{figure}[h]
		\centering
		\begin{tikzpicture}
    \node (wa) [wa]  {Invariant Extended Kalman Filter};
    \path (wa.west)++(-2.8,0.0) node[text width=0.7cm,align=right] (asr2) {$\boomega_n^{\mathrm{\textsc{imu}}}$  $\bfa_n^{\mathrm{\textsc{imu}}}$};
       
    \path (wa.east)+(0.6,0) node[right, text width=0.8cm] (vote)  {$\hbfx_{n+1}$, $\bfP_{n+1}$};

    \path [draw, ->] (asr2.east) -- (wa.west);
    \path [draw, ->] (wa.east) -- (vote.west);
        
    \path (wa.north)+(-1.5,1.0) node[de] (de)  {AI-based Noise Parameter Adapter};
    \path (wa.south) +(-1,-0.5) node (asrs) {proposed IMU dead-reckoning system};
  
  	 \path [draw, ->] (asr2.east) -- ++(0.5,0) |-  (de.west);
  	 
  	 \path [draw, ->] (de.east) -- ++(0.5,0) node [above] {$\bfN_{n+1}$} -- ++(0,-1);
  
    \begin{pgfonlayer}{background}
        \path (wa.south)+(-4,-1) node (a) {};
        \path (de.north)+(3.75,0.5) node (c) {};
        \path[rounded corners, draw=black!50, dashed]
            (a) rectangle (c);           
    \end{pgfonlayer}
\end{tikzpicture}
		\caption{Structure of the proposed system for inertial dead-reckoning. The measurement noise adapter feeds the filter with measurement covariance from raw IMU signals only.  \label{fig:scheme}}
	\end{figure}
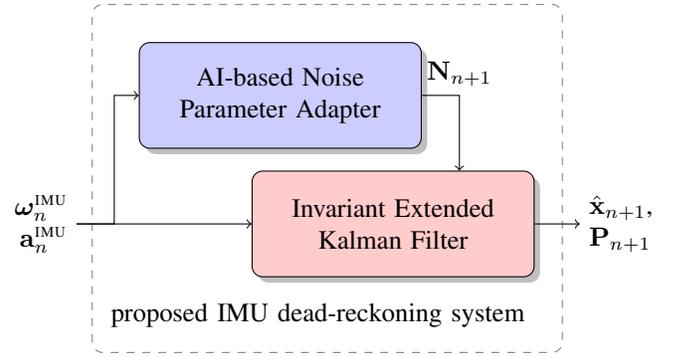

	This section describes our system for recovering all the variables of interest  from   IMU signals only. Figure \ref{fig:scheme} illustrates the approach which consists of two main blocks summarized as follows: 
	\begin{itemize}
		\item the filter integrates the inertial measurements \eqref{eq:gyro}-\eqref{eq:acc} with dynamical model $f(\cdot)$ given by \eqref{eq:propend2}-\eqref{eq:prop3} and \eqref{eq:ass3}-\eqref{eq:ass4}, and exploits  \eqref{eq:ass2} as measurements $h(\cdot)$ with covariance matrix $\bfN_{n}$ to refine its estimates;
		\item the noise parameter adapter determines in real-time the most suitable covariance noise matrix $\bfN_{n}$. This deep learning based adapter converts directly raw IMU signals \eqref{eq:gyro}-\eqref{eq:acc} into covariance matrices $\bfN_{n}$ without requiring knowledge of any state estimate nor any other quantity.
	\end{itemize}
	The amplitude of process noise parameters $\bfQ_n$ are considered fixed by the algorithm,  and are learned during training. 
	
	Note that the adapter computes covariances meant to improve localization  accuracy, and thus the computed values may  broadly differ from the actual statistical covariance  of $\bfy_n$  \eqref{eq:ass2}, see Section \ref{sec:covnet_eval} for more details. In this respect, our approach is  related to \cite{haarnojaBackprop2016} but the considered problem  is   more challenging:   our   state-space is of dimension 21  whereas \cite{haarnojaBackprop2016} has a state-space of dimension   only 3. Moreover, we   compare our results  in the sequel to state-of-the-art methods based on stereo cameras and LiDARs, and we show we may achieve similar results based on the moderately precise  IMU only.

	\subsection{AI-based Measurement Noise Parameter Adapter}\label{sec:covnet}

	The measurement noise parameter adapter computes at each instant $n$ the covariance $\bfN_{n+1}$ used in the filter update, see Figure \ref{fig:filter}. The base core of the adapter is a Convolutional Neural Network (CNN) \cite{goodfellowDeep2016}. The networks takes as input a window of $N$ inertial measurements and computes
	\begin{align}
	\bfN_{n+1} &= \mathrm{CNN}\left(\left\lbrace\boomega^{\textsc{imu}}_i,~ \bfa^{\textsc{imu}}_i\right\rbrace_{i=n-N}^n \right). \label{eq:cnn}
	\end{align}
	Our motivations for  the above simple CNN-like architecture are threefold: \begin{enumerate}
		\item[$i$)] avoiding over-fitting by using a relatively small number of parameters in the network  and also by making its outputs   independent of state estimates; 
		\item[$ii$)] obtaining an interpretable adapter from which one can infer general and safe rules using reverse engineering, e.g. to which extent must one inflate the covariance during turns,  for e.g.,  generalization to all sorts of wheeled and commercial vehicles, see Section \ref{sec:covnet_eval}; 
		\item[$iii$)]  letting the network be trainable. Indeed, as reported in  \cite{haarnojaBackprop2016}, training is quite difficult and slow. Setting the adapter with a recurrent architecture \cite{goodfellowDeep2016} would make the training even much harder.
	\end{enumerate}
	
	The complete architecture of the adapter consists of a bunch of CNN layers followed by a full layer outputting a vector $\bfz_n= \left[z_n^{\mathrm{lat}},z_n^{\mathrm{up}}\right]^T \in \bbR^2$. The covariance $\bfN_{n+1} \in\mathbb R^{2\times 2}$ is then computed as 
	\begin{align}
	\bfN_{n+1} = \diag\left( \sigma_{\mathrm{lat}}^2 10^{\beta \tanh(z_n^{\mathrm{lat}})},~ \sigma_{\mathrm{up}}^2 10^{\beta\tanh(z_n^{\mathrm{up}})} \right), \label{eq:N}
	\end{align} 
	with $\beta \in \bbR_{>0}$, and  where $\sigma_{\mathrm{lat}}$ and $\sigma_{\mathrm{up}} $ correspond to our initial guess for the noise parameters. The network thus may inflate covariance up to a factor $10^{\beta}$ and squeeze it up to a factor $10^{-\beta}$ with respect to its original values. Additionally, as long as the network is disabled or barely reactive (e.g. when starting training), we get $\bfz_n \approx \bfzero$ and recover the initial  covariance $\diag\left(\sigma_{\mathrm{lat}},~ \sigma_{\mathrm{up}} \right)^2$.
	
	Regarding process noise parameter $\bfQ_n$, we choose to fix it to a value $\bfQ$ and leave its dynamic adaptation for future work. However its entries are optimized during training, see Section \ref{sec:training}.

	\subsection{Implementation Details}\label{sec:implementation}
	We provide in this section the setting and the implementation details of our method. We implement the full approach in Python with the PyTorch\footnote{\url{https://pytorch.org/}} library for the noise parameter adapter part. We set as initial values before training
	\begin{align}
	\bfP_0 &= \diag\left(\sigma_0^{\bfR} \bfI_{2}, 0, \sigma_0^{\bfv} \bfI_{2}, \bfzero_4, \sigma_0^{\bfb^{\boomega}} \bfI, \sigma_0^{\bfb^{\bfa}} \bfI, \sigma_0^{\bfR^\mathsf{c}} \bfI,  \sigma_0^{\bfp^\mathsf{c}} \bfI \right)^2, \label{eq:P_0}\\
	\bfQ &= \diag\left(\sigma_{\boomega}\bfI,~\sigma_{\bfa}\bfI,~ \sigma_{\bfb^{\boomega}}\bfI,~ \sigma_{\bfb^{\bfa}}\bfI,~ \sigma_{\bfR^\mathsf{c}}\bfI,~ \sigma_{\bfp^\mathsf{c}}\bfI\right)^2, \label{eq:Q_n}\\
	\bfN_n &= \diag\left(\sigma_{\mathrm{lat}},~ \sigma_{\mathrm{up}} \right)^2, \label{eq:N_n}
	\end{align}
	where $\bfI = \bfI_3$, $\sigma_0^{\bfR} = \SI{e-3}{rad}$, $\sigma_0^{\bfv} = \SI{0.3}{m/s}$, $\sigma_0^{\bfb^{\boomega}} = \SI{e-4}{rad/s}$, $\sigma_0^{\bfb^{\bfa}} = \SI{3e-2}{m/s^2}$, $\sigma_0^{\bfR^\mathsf{c}} = \SI{3e-3}{rad}$, $\sigma_0^{\bfp^\mathsf{c}} = \SI{e-1}{m}$ in the initial error covariance $\bfP_0$, $\sigma_{\boomega} = \SI{1.4e-2}{rad/s}$,  $\sigma_{\bfa} = \SI{3e-2}{m/s^2}$, $\sigma_{\bfb^{\boomega}} = \SI{e-4}{rad/s}$, $\sigma_{\bfb^{\bfa}} = \SI{e-3}{m/s^2}$, $\sigma_{\bfR^\mathsf{c}} = \SI{e-4}{rad}$, $\sigma_{\bfp^\mathsf{c}} = \SI{e-4}{m}$ for the noise propagation covariance matrix $\bfQ$,  $\sigma_{\mathrm{lat}} = \SI{1}{m/s}$,  and $\sigma_{\mathrm{up}} = \SI{3}{m/s}$ for the measurement covariance matrix. The zero values in the diagonal of $\bfP_0$ in \eqref{eq:P_0} corresponds to a perfect prior of the initial yaw, position and zero vertical speed.
	
	The adapter is a 1D temporal convolutional neural network with 2 layers. The first layer has kernel size 5, output dimension 32, and dilatation parameter 1. The second has kernel size 5, output dimension 32 and dilatation parameter 3, thus it set the window size equal to $N=15$. The CNN is followed by a fully connected layer that output the scalars $z^{\mathrm{lat}}$ and $z^{\mathrm{up}}$. Each activation function between two layers is a ReLU unit \cite{goodfellowDeep2016}.
	We define $\beta = 3$ in the right part of \eqref{eq:N}  which allows for each covariance element to be $10^3$ higher or smaller than its original values.

	\subsection{Training}\label{sec:training}
	We seek to optimize the relative translation error $t_{rel}$ computed from the filter estimates $\hbfx_n$, which is the averaged increment error for all possible sub-sequences of length \SI{100}{m} to \SI{800}{m}. 
	
	Toward this aim, we first define the learnable parameters. It consists of the 6210 parameters of the adapter, along with the parameter elements of $\bfP_0$ and $\bfQ$ in \eqref{eq:P_0}-\eqref{eq:Q_n}, which add 12 parameters to learn. We then choose an Adam optimizer \cite{kingmaAdam2014} with learning rate $10^{-4}$ that updates the trainable parameters. Training consists of repeating for a chosen number of epochs the following iterations:
	\begin{enumerate}
		\item[$i$)] sample a part of the dataset;
		\item[$ii$)] get the filter estimates for then computing loss and gradient w.r.t. the learnable parameters;
		\item[$ii$)] update the learnable parameters with gradient and optimizer.
	\end{enumerate}
	Following continual training \cite{parisiContinual2019}, we suppose the number of epochs is huge, potentially infinite (in our application we set this number to 400). This makes sense for online training in a context where the vehicle gathers accurate ground-truth poses from e.g. its LiDAR system or precise GNSS. It requires careful procedures for avoiding over-fitting, such that we use dropout and data augmentation \cite{goodfellowDeep2016}. Dropout refers to ignoring units of the adapter during training, and we set the probability $p=0.5$ of any CNN element to be ignored (set to zero) during a sequence iteration.
	
	\renewcommand{\figurename}{Table}
	\setcounter{figure}{0} 
	\begin{figure*}[t]
		\centering
		\begin{tabular}{c||c|c|c||c|c||c|c||c|c||c|c}
			\toprule
			\multirow{3}{0.5cm}{test seq.}  &    &  & \multirow{3}{*}{environment} & \multicolumn{2}{c|}{IMLS \cite{deschaudIMLSSLAM2018}} & \multicolumn{2}{c|}{ORB-SLAM2 \cite{mur-artalORBSLAM22017}} & \multicolumn{2}{c|}{IMU} &  \multicolumn{2}{c}{\textbf{proposed}}   \\	
			& length  & duration &  & $t_{rel}$ & $r_{rel}$ & $t_{rel}$ & $r_{rel}$ & $t_{rel}$ & $r_{rel}$ & $t_{rel}$ & $r_{rel}$ \\
			& (\si{km}) & (\si{s}) &  &(\%) & (\si{deg/m})& (\%)& (\si{deg/m})& (\%) & (\si{deg/m})& (\%) & (\si{deg/m})\\
			\midrule
			01 & 2.6 & 110 & highway & \textbf{0.48} & \textbf{0.08} & 1.38 & 0.20 & 5.35 & 0.12 &  {1.11} &  {0.12} \\
			03 & - & 80 & country & - & - & - & - & - & - & - & - \\
			04 & 0.4 & 27 & country & \textbf{0.25} & \textbf{0.08} & 0.41 & 0.21  & 0.97 & 0.10 &   {0.35} & \textbf{0.08} \\
			06 & 1.2 & 110 & urban & \textbf{0.78} & \textbf{0.07} &   {0.89} & 0.22  & 5.78 &   {0.19} & 0.97 & 0.20 \\
			07 & 0.7 & 110 & urban & \textbf{0.32} & \textbf{0.12} & 1.16 & 0.49 & 12.6 &   {0.30} &   {0.84} & 0.32 \\
			08 & 3.2 & 407 & urban, country & 1.84 & \textbf{0.17} &   {1.52} &   {0.30} & 549 & 0.56 & \textbf{1.48} & 0.32 \\
			09 & 1.7 & 159 & urban, country &   {0.97} & \textbf{0.11} & 1.01 & 0.25 & 23.4 & 0.32 & \textbf{0.80} &   {0.22}\\
			10 & 0.9 & 120 & urban, country &   {0.50} & \textbf{0.14} & \textbf{0.31} & 0.34 & 4.58 & 0.25 & 0.98 &   {0.23} \\ \midrule[1.4pt]
			\multicolumn{4}{c|}{average scores} & \textbf{0.98} & \textbf{0.12} & 1.17 & 0.27 & 171 & 0.31 & 1.10 & 0.23 \\
			\bottomrule
		\end{tabular}
		\caption{Results on \cite{geigerVision2013}.  IMU integration tends to drift or diverge, whereas the proposed method may be used as an alternative to LiDAR based (IMLS) and stereo vision based (ORB-SLAM2) methods, using only IMU information. Indeed, on average, our dead-reckoning solution performs better than ORB-SLAM2 and achieves a  translational error being close to that of the LiDAR based method IMLS, which is ranked 3\textsuperscript{rd}  on the KITTI online benchmarking system.  Data from seq. \texttt{03} was unavailable for testing algorithms, and sequences \texttt{00}, \texttt{02} and \texttt{05} are  discussed separately in Section \ref{sec:results2}. It should be duly noted IMLS, ORB-SLAM2, and the proposed AI-IMU algorithm, all use different sensors. The interest of ranking algorithms based on different information is debatable. Our goal here is rather to evidence that using data from a moderately precise IMU only, one can achieve similar results as state of the art systems based on imaging sensors, which is a rather surprising feature.  \label{fig:table}}
	\end{figure*}
	\renewcommand{\figurename}{Fig.}
	\setcounter{figure}{4} 
	
	Regarding $i$), we sample a batch of nine \SI{1}{min} sequences, where each sequence starts at a random arbitrary time. We add to data a small Gaussian noise with standard deviation $10^{-4}$, a.k.a. data augmentation technique. We compute $ii$) with standard backpropagation, and we finally clip the gradient norm to a maximal value of $1$ to avoid gradient explosion at step $iii$). 
	
	We stress the loss function consists of the relative translation error $t_{rel}$, i.e. we optimize parameters for improving the filter accuracy, disregarding  the values actually  taken by  $\bfN_n$, in the spirit of \cite{haarnojaBackprop2016}.

	\section{Experimental Results}\label{sec:results}

	We evaluate the proposed method on the KITTI dataset \cite{geigerVision2013}, which contains data recorded from LiDAR, cameras and IMU, along with centimeter accurate ground-truth pose from different environments (e.g., urban, highways, and streets).  The dataset contains 22 sequences for benchmarking odometry algorithms, 11 of them contain publicly available ground-truth trajectory, raw and synchronized IMU data. We download the raw data with IMU signals sampled at \SI{100}{Hz} ($dt=\SI{e-2}{s}$) rather than the synchronized data sampled at \SI{10}{Hz}, and discard seq. \texttt{03} since we did not find raw data for this sequence. The RT3003\footnote{\url{https://www.oxts.com/}} IMU has announced gyro and accelerometer bias stability of respectively \SI{36}{\deg/h} and \SI{1}{mg}. The KITTI dataset has an online benchmarking   system that ranks algorithms. However we could not submit our algorithm for online ranking since sequences used for ranking do not contain IMU data, which is reserved for training only. Our implementation is made open-source at:
	\begin{center}\texttt{\url{https://github.com/mbrossar/ai-imu-dr}}. \end{center}

	\subsection{Evaluation Metrics and Compared Methods}\label{sec:metric}
	To assess performances we consider the two error metrics proposed in \cite{geigerVision2013}: 
	\subsubsection{Relative Translation Error ($t_{rel}$)} which is the averaged relative translation increment error for all possible sub-sequences of length \SI{100}{m}, \ldots, \SI{800}{m}, in percent of the traveled distance;
	\subsubsection{Relative Rotational Error ($t_{rel}$)} that is the relative rotational increment error for all possible sub-sequences of length \SI{100}{m}, \ldots, \SI{800}{m}, in degree per meter.
	
	We compare four methods which alternatively use  LiDAR, stereo vision, and IMU-based estimations:
	\begin{itemize}
		\item \textbf{IMLS} \cite{deschaudIMLSSLAM2018}: a recent state-of-the-art LiDAR-based approach ranked 3\textsuperscript{rd}  in the KITTI benchmark.  The author provided us with the code after disabling the loop-closure module;
		\item \textbf{ORB-SLAM2} \cite{mur-artalORBSLAM22017}: a popular and versatile library for monocular, stereo and RGB-D cameras that computes a sparse reconstruction of the map. We took the open-source code, disable loop-closure capability and evaluate the stereo algorithm without modifying any parameter;
		\item \textbf{IMU}: the direct integration of the IMU measurements based on \eqref{eq:propend}-\eqref{eq:prop1}, that is, pure inertial navigation;
		\item \textbf{proposed}: the proposed approach, that uses only the IMU signals and involves no other sensor. 
	\end{itemize}

	\subsection{Trajectory Results}\label{sec:traj}
	
	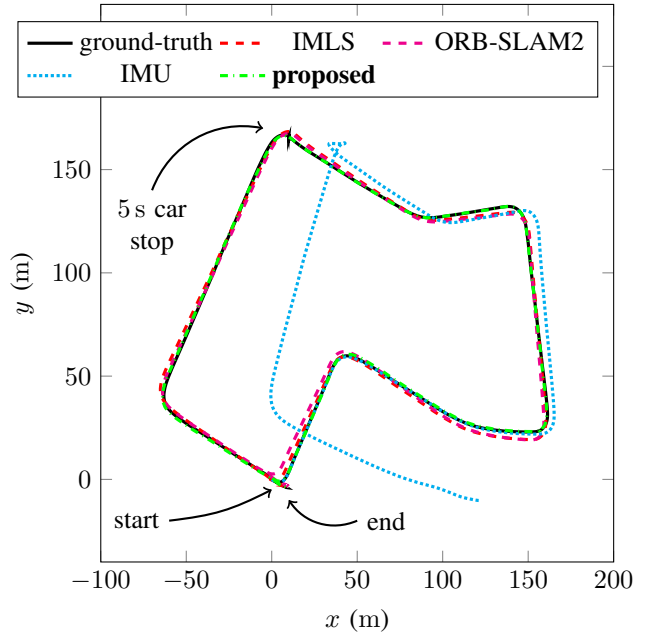
\begin{figure}
		\centering
		\begin{tikzpicture}[]
		\begin{axis}[height=9cm,
		width=8.4cm,
		ylabel=$y$ (\SI{}{m}),
		xlabel=$x$ (\SI{}{m}),
		xmax=200,
		xmin=-100,
		ymax=230,
		ymin=-40,
		legend columns=3, legend pos= north east, 
		ticks=both,
		legend entries={ground-truth, IMLS, ORB-SLAM2, IMU, \textbf{proposed}}
		]
		
		\addplot[draw = black, solid,smooth,
		very thick] table[x=x,y=y] {07gt.txt};
		\addplot[draw = red, smooth,solid, dashed, very thick] table[x=x,y=y] {07p_imls.txt};
		\addplot[draw = magenta, smooth, dashed, very thick] table[x=x,y=y] {07p_orb.txt};
		\addplot[draw = cyan, smooth, densely dotted, very thick] table[x=x,y=y] {07p_imu.txt};
		\addplot[draw = green, smooth, dashdotted, very thick] table[x=x,y=y] {07p_rinsw.txt};

		\path [draw, thick, ->] (50, -20)  node [right] {end} to[bend  right=-40,looseness=1.0, ->] (10, -10);
		\path [draw, thick, ->] (-60, -20)  node [left] {start} to[bend  right=10,looseness=1.1, ->] (-0, -5);
		\path [draw, thick, ->] (-70, 140)  node [below, text centered, text width=1cm] {\SI{5}{s} car stop} to[bend  right=-40,looseness=1, ->] (-5, 170);
		\end{axis}	
		\end{tikzpicture}
		\caption{Results on seq. \texttt{07} (drive \#27, 2011/09/30) \cite{geigerVision2013}. The proposed method competes with LiDAR and visual odometry methods, whereas the IMU integration broadly drifts after the car stops.\label{fig:07}} 
	\end{figure}
	
	\begin{figure}
		\centering
		\begin{tikzpicture}[]
		\begin{axis}[height=9.2cm,
		width=8.1cm,
		ylabel=$y$ (\SI{}{m}),
		xlabel=$x$ (\SI{}{m}),
		xmax=550,
		xmin=-100,
		ymax=220,
		ymin=-580,
		legend columns=3, legend pos= north east, 
		ticks=both,
		legend entries={ground-truth, IMLS, ORB-SLAM2, IMU, \textbf{proposed}},
		]
		
		\addplot[draw = black, solid,smooth,
very thick] table[x=x,y=y] {09gt.txt};
\addplot[draw = red, smooth,solid, dashed,very thick] table[x=x,y=y] {09p_imls.txt};
\addplot[draw = magenta, smooth, dashed,very thick] table[x=x,y=y] {09p_orb.txt};
\addplot[draw = cyan, smooth, densely dotted,very thick] table[x=x,y=y] {09p_imu.txt};
\addplot[draw = green, smooth, dashdotted,very thick] table[x=x,y=y] {09p_rinsw.txt};
		\path [draw, thick, ->] (70, -60)  node [right] {start/end} to[bend  right=-30,looseness=1., ->] (-5, -10);

		\end{axis}	
		\end{tikzpicture}
		
		\caption{Results on seq. \texttt{09} (drive \#33, 2011/09/30) \cite{geigerVision2013}. The proposed method competes with LiDAR and visual odometry methods, whereas the IMU integration drifts  quickly after the first turn. \label{fig:09}} 
	\end{figure}
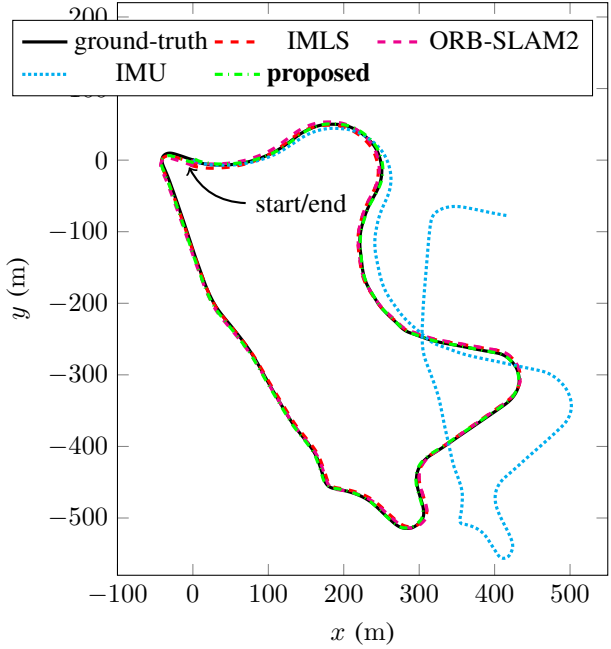

	We follow the same protocol for evaluating each sequence: $i$) we initialize the filter with parameters described in Section \ref{sec:implementation}; $ii$) we train then the noise parameter adapter following  Section \ref{sec:training} for 400 epochs without the evaluated sequence (e.g. for testing seq. \texttt{10}, we train on seq. \texttt{00}-\texttt{09})  so that the noise parameter has \emph{never} been confronted with  the evaluated sequence; $iii$) we run the IMU-based methods on the full raw sequence with ground-truth initial configuration ($\bfR_0^{\textsc{imu}}, \bfv_0^{\textsc{imu}}, \bfp_0^{\textsc{imu}}$), whereas we initialize remaining variables at zero ($\bfb_0^{\boomega}= \bfb_0^{\bfa}=\bfp_0^{\mathsf{c}}=\bfzero$, $\bfR_0^{\mathsf{c}}=\bfI$); and $iv$) we get the estimates only on time corresponding to the odometry benchmark sequence. LiDAR and visual methods are directly evaluated on the odometry sequences.

	Results are averaged in Table \ref{fig:table} and illustrated in Figures \ref{fig:example}, \ref{fig:07} and \ref{fig:09}, where we exclude sequences \texttt{00}, \texttt{02} and \texttt{05} which contain problems with the data, and will be  discussed separately in Section \ref{sec:results2}. From these results, we see that:
	\begin{itemize}
		\item LiDAR and visual methods perform  generally well in all sequences, and the LiDAR method achieves slightly better results than its visual counterpart;
		\item our method competes on average with the latter image based methods,  see Table \ref{fig:table};
		\item directly integrating the IMU signals leads to rapid drift of the estimates, especially for the longest sequences but even for short periods;
		\item Our method  looks unaffected by stops of the car, as in seq. \texttt{07}, see Figure \ref{fig:07}. 
	\end{itemize}
	The results are remarkable as we use none of the vision sensors, nor wheel odometry. We only use the IMU, which moreover has moderate precision.

	We also sought to compare our method to visual inertial odometry algorithms. However, we could not find open-source of such method that performs well on the full KITTI dataset. We tested \cite{qinGeneral2019} but the code in still under development (results sometimes diverge), and the authors in \cite{ramezaniVehicle2018} evaluate their not open-source  method for short sequences ($\leq$ \SI{30}{s}). The paper \cite{heoEKFBased2018, heoConsistent2018} evaluate their visual inertial odometry methods on seq. \texttt{08}, both get a final error around \SI{20}{m}, which is four times what  our method gets, with final distance to ground-truth of  only \SI{5}{m}. This clearly evidences that methods taylored for ground vehicles \cite{wuVINS2017,zhengSE2018} may achieve  higher accuracy and robustness that general methods designed for smartphones, drones and aerial vehicles.

	\subsection{Results on Sequences \texttt{00}, \texttt{02} and \texttt{05}} \label{sec:results2}

	Following the procedure described in Section \ref{sec:traj}, the proposed method seems to have degraded performances on seq. \texttt{00}, \texttt{02} and \texttt{05}, see e.g. Figure \ref{fig:02}. However, the behavior is wholly explainable: data are missing for a couple of seconds due to logging problems which appear both for IMU and ground-truth. This is  illustrated  in Figure \ref{fig:data10} for seq. \texttt{02} where we plot available data over time. We observe jump in the IMU and ground-truth signals, that illustrate  data are missing between $t=1$ and $t=3$.  The problem was corrected manually when using those sequences in the training phase described in Section \ref{sec:results}.

	Although those sequences could have been discarded due to logging problems, we used them for testing without correcting their problems. This naturally  results in degraded performance, but also evidences  our method is remarkably robust to such problems in spite of their inherent harmfulness. For instance, the \SI{2}{s} time jump of seq. \texttt{02} results in   estimate shift, but no divergence occurs for all that, see Figure \ref{fig:02}.

	\begin{figure}
		\centering
		\begin{tikzpicture}[]
		\begin{axis}[height=7.6cm,
		width=8.1cm,
		ylabel=$y$ (\SI{}{m}),
		xlabel=$x$ (\SI{}{m}),
		xmax=1250,
		xmin=950,
		ymax=-1500,
		ymin=-1850,
		ylabel style={xshift=-0.0cm,yshift=-0.0cm},
		xlabel style={xshift=0cm,yshift=0cm},
		legend columns=2, legend pos= north east,  
		ticks=both,
		legend entries={ground-truth, prop. w/o alignment, \textbf{proposed}, prop. w/o cov. adapter},
		]
		
		\addplot[draw = black, 
		very thick] table[x=y,y=x] {01gt.txt};
		\addplot[draw = red, dashed,very thick] table[x=y,y=x] {01p_rinsw.txt};
		\addplot[draw = green, dashdotted,very thick] table[x=y,y=x] {01p_rinsw_sans_deep.txt};
		\addplot[draw = orange, dashed,very thick] table[x=y,y=x] {01p_rinsw_sans_alignment.txt};
		
		\end{axis}	
		\end{tikzpicture}
		\caption{End trajectory results on the highway seq. \texttt{01} (drive \#42, 2011/10/30) \cite{geigerVision2013}. Dynamically adapting the measurement covariance and considering misalignment between car and inertial frames enhance the performances of the proposed method from a translational error of 1.94\% to one of 1.11\%. \label{fig:01}} 
	\end{figure}
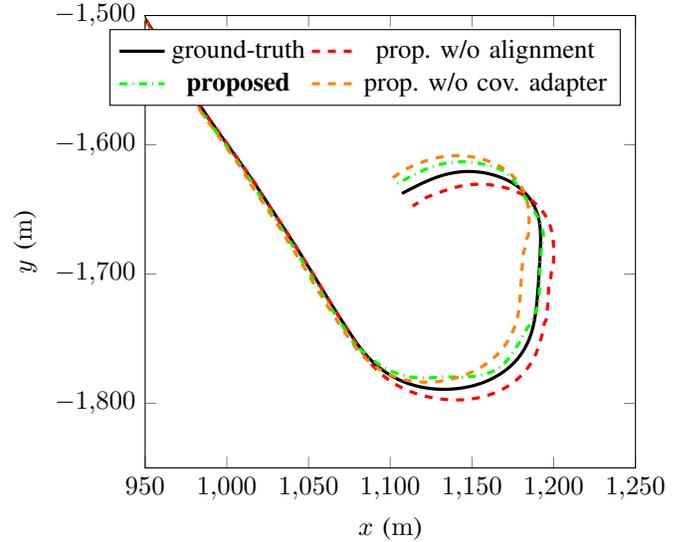
	
	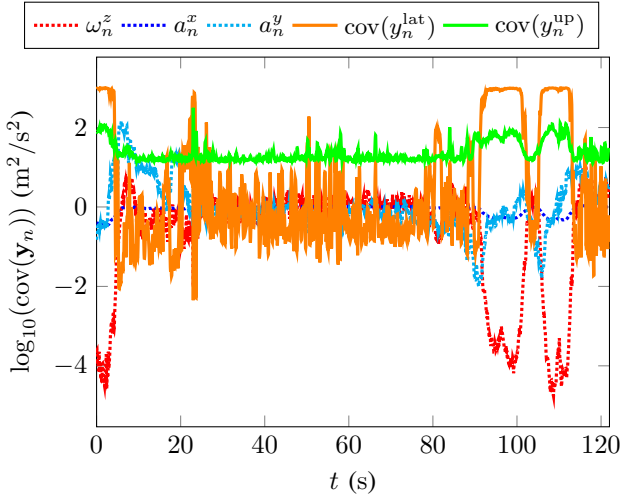
\begin{figure}
		\centering
		\begin{tikzpicture}[]
		\begin{axis}[height=6.5cm,
		width=8.4cm,
		ylabel=$\log_{10}(\cov(\bfy_n))$ (\si{m^2/s^2}),
		xlabel=$t$ (\SI{}{s}),
		xmax=122,
		xmin=0,
		ylabel style={xshift=-0.0cm,yshift=-0.0cm},
		xlabel style={xshift=0cm,yshift=0cm},
		legend columns=5,
		legend style={at={(0.98,1.15)}},
		ticks=both,
		legend entries={$\omega^{z}_{n}$, $a^{x}_{n}$, $a^{y}_{n}$, $\cov(y_n^{\mathrm{lat}})$, $\cov(y_n^{\mathrm{up}})$},
		legend style={font=\small}
		]
		
		\addplot[draw=red, smooth, densely dotted,very thick] table[x=t,y=accy] {01data.txt};
		\addplot[draw=blue, smooth, densely dotted,very thick] table[x=t,y=omegaz] {01data.txt};
		\addplot[draw=cyan, smooth, densely dotted,very thick] table[x=t,y=accx] {01data.txt};
		\addplot[draw = orange, smooth,very thick] table[x=t,y=covlat] {01data.txt};
		\addplot[draw = green, smooth,very thick] table[x=t,y=covup] {01data.txt};
		
		\end{axis}	
ea		\end{tikzpicture}
		\caption{Covariance values computed by the adapter on the highway seq. \texttt{01} (drive \#42, 2011/10/30) \cite{geigerVision2013}. We clearly observe a large increase in the covariance values when the car is turning between $t=\SI{90}{s}$ and $t=\SI{110}{s}$. \label{fig:cov01}} 
	\end{figure}

	\subsection{Discussion}\label{sec:covnet_eval}
	The performances are owed to three components: $i$) the use of a recent IEKF that has been proved to be well suited for IMU based localization; $ii$) incorporation of side  information in the form of  pseudo-measurements with dynamic noise parameter adaptation learned by a neural network; and $iii$)  accounting for  a ``loose" misalignment between the IMU frame and the car frame. 
	
	As concerns $i$), it should be stressed the method is perfectly suited to the use of a conventional EKF and is easily adapted if need be. However we advocate the use of an IEKF owing to its accuracy and convergence properties. To illustrate the benefits of points $ii$) and $iii$), we consider two sub-versions of the proposed algorithm. One without alignment, i.e. where $\bfR_n^{\mathsf{c}}$ and $\bfp_n^{\mathsf{c}}$ are not included in the state and fixed at   their initial values $\bfR^{\mathsf{c}}_n=\bfI$, $\bfp^{\mathsf{c}}_n=\bfzero$, and a second one that uses the static filter parameters \eqref{eq:P_0}-\eqref{eq:N_n}.
	
	End trajectory results for the highway seq. \texttt{01} are plotted in Figure \ref{fig:01}, where we see that the two sub-version methods have trouble when the car is turning. Therefore their respective translational errors $t_{rel}$ are higher than the full version of the proposed method:  the proposed method achieves 1.11\%, the method without alignment level arm achieves  1.65\%, and the absence of covariance adaptation yields 1.94\% error. All methods have the same rotational error $r_{rel}=\SI{0.12}{\deg/m}$. This could be anticipated for the considered sequence since the full method has the same  rotational error than standard IMU integration method.
	
	As systems with equipped  AI-based approaches may be hard to certify for  commercial or industrial use \cite{sunderhauflimits2018}, we note adaptation rules may be inferred from the AI-based adapter, and encoded in an EKF using pseudo-measurements. To this aim, we plot the covariances computed by the adapter for seq. \texttt{01}  in Figure \ref{fig:cov01}. The adapter clearly increases the covariances during the bend, i.e. when the gyro yaw rate is important. This is especially the case for the zero velocity measurement \eqref{eq:ass2}: its associated covariance is inflated by a factor of $10^2$ between $t=\SI{90}{s}$ and $t=\SI{110}{s}$. This illustrates the kind of   information the  adapter has learned. Interestingly, we see large discrepancies may occur between the actual statistical uncertainty (which should clearly be below  \SI{100}{m^2/s^2})  and the inflated covariances whose values are computed for the sole purpose of   filter's performance enhancement. Indeed, such a large noise parameter inflation indicates the AI-based part of the algorithm has learned and recognizes that pseudo-measurements have no value for localization at those precise  moments, so the filter should barely consider them.

	\section{Conclusion}\label{sec:con}
	This paper proposes a novel approach for inertial dead-reckoning for wheeled vehicles. Our approach exploits deep neural networks to dynamically adapt the covariance of  simple assumptions about the vehicle motions which are leveraged in an invariant extended Kalman filter that performs localization, velocity and sensor bias estimation. The entire algorithm is fed with IMU signals only, and requires no other sensor. The method leads to surprisingly accurate results, and opens  new perspectives. In future work, we would like to address the   learning of the Kalman covariance matrices for images, and also the issue of generalization from one vehicle to another.

	\section*{Acknowledgments}
	The authors would like to thank J-E. \textsc{Deschaud} for sharing the results of the IMLS algorithm \cite{deschaudIMLSSLAM2018}, and Paul \textsc{Chauchat} for relevant discussions. 
	
	\renewcommand{\figurename}{Table}
	\setcounter{figure}{1} 
	\begin{figure*}
		\centering
		\begin{tabular}{c||c|c|c||c|c|c|c|c|c|c|c}
			\toprule
			\multirow{3}{0.5cm}{test seq.}  &    &  & \multirow{3}{*}{environment} & \multicolumn{2}{c|}{IMLS \cite{deschaudIMLSSLAM2018}} & \multicolumn{2}{c|}{ORB-SLAM2 \cite{mur-artalORBSLAM22017}} & \multicolumn{2}{c|}{IMU} &  \multicolumn{2}{c}{\textbf{proposed}}   \\	
			& length  & duration &  & $t_{rel}$ & $r_{rel}$ & $t_{rel}$ & $r_{rel}$ & $t_{rel}$ & $r_{rel}$ & $t_{rel}$ & $r_{rel}$ \\
			& (\si{km}) & (\si{s}) &  &(\%) & (\si{deg/m})& (\%)& (\si{deg/m})& (\%) & (\si{deg/m})& (\%) & (\si{deg/m})\\
			\midrule
			00 & 3.7 &  454 & urban & 1.46 & \textbf{0.17} & \textbf{1.33} & 0.29 & 426 & 4.68 & 7.40 & 2.65 \\ 
			02 & 5.1 & 466 & urban & 2.42 & \textbf{0.12} & \textbf{1.84} & 0.28  & 346 & 0.87 & 2.85 & 0.38\\
			05 & 2.2 & 278 & urban & 0.83 & \textbf{0.10} & \textbf{0.75} & 0.22  & 189 & 0.52 & 2.76 & 0.86  \\
			\bottomrule
		\end{tabular}
		\caption{Results on \cite{geigerVision2013} on seq. \texttt{00}, \texttt{02} and \texttt{05}. The degraded results of the proposed method are wholly explained by a problem of missing data, see Section \ref{sec:results2} and Figure \ref{fig:data10}. \label{fig:table2}}
	\end{figure*}
	\renewcommand{\figurename}{Fig.}
	\setcounter{figure}{8} 
	
	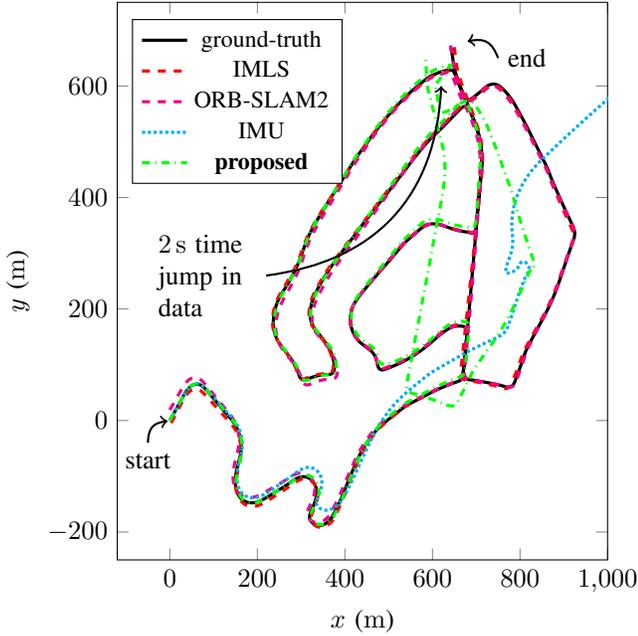
\begin{figure}
		\centering
		\begin{tikzpicture}[]
		\begin{axis}[height=9cm,
		width=8.1cm,
		ylabel=$y$ (\SI{}{m}),
		xlabel=$x$ (\SI{}{m}),
		xmax=1000,
		xmin=-120,
		ymax=750,
		ymin=-250,
		legend columns=1, legend pos= north west, 
		ticks=both,
		legend entries={ground-truth, IMLS, ORB-SLAM2, IMU, \textbf{proposed}},
		legend style={font=\small},
		]
		
		\addplot[draw = black, solid,smooth,
very thick] table[x=x,y=y] {02gt.txt};
\addplot[draw = red, smooth,solid, dashed,very thick] table[x=x,y=y] {02p_imls.txt};
\addplot[draw = magenta, smooth, dashed,very thick] table[x=x,y=y] {02p_orb.txt};
\addplot[draw = cyan, smooth, densely dotted,very thick] table[x=x,y=y] {02p_imu.txt};
\addplot[draw = green, smooth, dashdotted,very thick] table[x=x,y=y] {02p.txt};

		\path [draw, thick, ->] (-50, -40)  node [below] {start} to[bend  right=-40,looseness=1.3, ->] (-5, -1);
		\path [draw, thick, ->] (220, 260)  node [left,text width=1.3cm] {\SI{2}{s} time jump in data} to[bend  right=40,looseness=1.1, ->] (620, 600);
		\path [draw, thick, ->] (750, 650)  node [right] {end} to[bend  right=40,looseness=1.3, ->] (680, 680);
		\end{axis}	
		\end{tikzpicture}
		
		\caption{Results on seq. \texttt{02} (drive 
			\#34, 2011/09/30) \cite{geigerVision2013}. The proposed method competes with LiDAR and visual odometry methods until a problem in data occurs (2 seconds are missing). It is remarkable that the proposed  method be robust to such trouble causing  a shift estimates but no  divergence.\label{fig:02}} 
	\end{figure}
	
	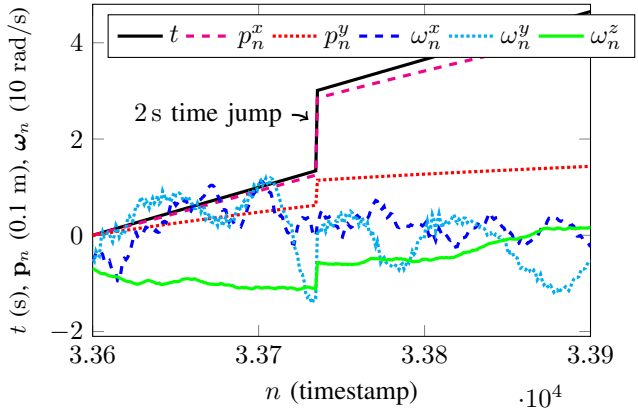
\begin{figure}
		\centering
		\begin{tikzpicture}[]
		\begin{axis}[height=6cm,
		width=8.2cm,
		ylabel={$t$ (\si{s}), $\bfp_{n}$ (0.1 \si{m}), $\boomega_{n}$ (10 \si{rad/s})},
		xlabel=$n$ (timestamp),
		xmax=33900,
		xmin=33600,
		ymax=4.8,
		ymin=-2.1,
		legend columns=6, legend pos= north west, 
		y label style={font=\small}, 
		ticks=both,
		legend entries={$t$, $p_{n}^x$, $p_{n}^y$, $\omega^{x}_{n}$, $\omega^{y}_{n}$, $\omega^{z}_{n}$},
		xtick={33600,33700, 33800,33900}
		]
		\addplot[draw = black,very thick] table[x=n,y=t] {02data.txt};
		\addplot[draw = magenta, dashed,very thick] table[x=n,y=x] {02data.txt};
		\addplot[draw = red, densely dotted, very thick] table[x=n,y=y] {02data.txt};
		\addplot[draw = blue,  dashed, very thick] table[x=n,y=omegax] {02data.txt};
		\addplot[draw = cyan, densely dotted, very thick] table[x=n,y=omegay] {02data.txt};
		\addplot[draw = green, very thick] table[x=n,y=omegaz] {02data.txt};
		\path [draw, thick, ->] (33720, 2.5)  node [left] {\SI{2}{s} time jump} to[bend  right=-20,looseness=1, ->] (33730, 2.4);
		\end{axis}	
		\end{tikzpicture}
		\caption{Data on seq. \texttt{02} (drive \#34, 2011/09/30) \cite{geigerVision2013}, as function of timestamps number $n$. A \SI{2}{s} time jump happen around $n=33750$, i.e. data have not been recorded during this jump. It leads to jump in ground-truth and IMU signals, causing estimate drift. \label{fig:data10}} 
	\end{figure}
	
	\bibliographystyle{IEEEtran}
	\bibliography{main}

\begin{thebibliography}{10}
\providecommand{\url}[1]{#1}
\csname url@samestyle\endcsname
\providecommand{\newblock}{\relax}
\providecommand{\bibinfo}[2]{#2}
\providecommand{\BIBentrySTDinterwordspacing}{\spaceskip=0pt\relax}
\providecommand{\BIBentryALTinterwordstretchfactor}{4}
\providecommand{\BIBentryALTinterwordspacing}{\spaceskip=\fontdimen2\font plus
\BIBentryALTinterwordstretchfactor\fontdimen3\font minus
  \fontdimen4\font\relax}
\providecommand{\BIBforeignlanguage}[2]{{%
\expandafter\ifx\csname l@#1\endcsname\relax
\typeout{** WARNING: IEEEtran.bst: No hyphenation pattern has been}%
\typeout{** loaded for the language `#1'. Using the pattern for}%
\typeout{** the default language instead.}%
\else
\language=\csname l@#1\endcsname
\fi
#2}}
\providecommand{\BIBdecl}{\relax}
\BIBdecl

\bibitem{bressonSimultaneous2017}
G.~Bresson, Z.~Alsayed, L.~Yu \emph{et~al.},
  ``\BIBforeignlanguage{en}{Simultaneous {{Localization}} and {{Mapping}}: {{A
  Survey}} of {{Current Trends}} in {{Autonomous Driving}}},''
  \emph{\BIBforeignlanguage{en}{IEEE Transactions on Intelligent Vehicles}},
  vol.~2, no.~3, pp. 194--220, 2017.

\bibitem{oxtsWhy2018}
{OxTS}, ``Why it is {{Necessary}} to {{Integrate}} an {{Inertial Measurement
  Unit}} with {{Imaging Systems}} on an {{Autonomous Vehicle}},'' https : / /
  www . oxts . com / technical-notes/why-use-ins-with-autonomous-vehicle/,
  2018.

\bibitem{geigerVision2013}
A.~Geiger, P.~Lenz, C.~Stiller \emph{et~al.}, ``\BIBforeignlanguage{en}{Vision
  meets robotics: {{The KITTI}} dataset},'' \emph{\BIBforeignlanguage{en}{The
  International Journal of Robotics Research}}, vol.~32, no.~11, pp.
  1231--1237, 2013.

\bibitem{barrauInvariant2017}
A.~Barrau and S.~Bonnabel, ``The {{Invariant Extended Kalman Filter}} as a
  {{Stable Observer}},'' \emph{IEEE Transactions on Automatic Control},
  vol.~62, no.~4, pp. 1797--1812, 2017.

\bibitem{barrauInvariant2018}
------, ``Invariant {{Kalman Filtering}},'' \emph{Annual Review of Control,
  Robotics, and Autonomous Systems}, vol.~1, no.~1, pp. 237--257, 2018.

\bibitem{deschaudIMLSSLAM2018}
J.-E. Deschaud, ``{{IMLS}}-{{SLAM}}: {{Scan}}-to-{{Model Matching Based}} on
  {{3D Data}},'' in \emph{International {{Conference}} on {{Robotics}} and
  {{Automation}}}.\hskip 1em plus 0.5em minus 0.4em\relax {IEEE}, 2018.

\bibitem{mur-artalORBSLAM22017}
R.~{Mur-Artal} and J.~Tardos, ``\BIBforeignlanguage{en}{{{ORB}}-{{SLAM2}}: {{An
  Open}}-{{Source SLAM System}} for {{Monocular}}, {{Stereo}}, and {{RGB}}-{{D
  Cameras}}},'' \emph{\BIBforeignlanguage{en}{Transactions on Robotics}},
  vol.~33, no.~5, pp. 1255--1262, 2017.

\bibitem{tschoppExperimental2019}
F.~{Tschopp}, T.~{Schneider}, A.~W. {Palmer} \emph{et~al.},
  ``\BIBforeignlanguage{en}{Experimental {{Comparison}} of {{Visual}}-{{Aided
  Odometry Methods}} for {{Rail Vehicles}}},''
  \emph{\BIBforeignlanguage{en}{IEEE Robotics and Automation Letters}}, vol.~4,
  no.~2, pp. 1815--1822, 2019.

\bibitem{mousaInertial2018}
M.~Mousa, K.~Sharma, and C.~G. Claudel, ``Inertial {{Measurement
  Units}}-{{Based Probe Vehicles}}: {{Automatic Calibration}}, {{Trajectory
  Estimation}}, and {{Context Detection}},'' \emph{IEEE Transactions on
  Intelligent Transportation Systems}, vol.~19, no.~10, pp. 3133--3143, 2018.

\bibitem{wahlstromMapAided2016}
J.~Wahlstrom, I.~Skog, J.~G.~P. Rodrigues \emph{et~al.},
  ``\BIBforeignlanguage{en}{Map-{{Aided Dead}}-{{Reckoning Using Only
  Measurements}} of {{Speed}}},'' \emph{\BIBforeignlanguage{en}{IEEE
  Transactions on Intelligent Vehicles}}, vol.~1, no.~3, pp. 244--253, 2016.

\bibitem{mahmoudIntegrated2019}
A.~Mahmoud, A.~Noureldin, and H.~S. Hassanein,
  ``\BIBforeignlanguage{en}{Integrated {{Positioning}} for {{Connected
  Vehicles}}},'' \emph{\BIBforeignlanguage{en}{IEEE Transactions on Intelligent
  Transportation Systems}}, pp. 1--13, 2019.

\bibitem{karlssonFuture2017}
R.~Karlsson and F.~Gustafsson, ``The {{Future}} of {{Automotive Localization
  Algorithms}}: {{Available}}, reliable, and scalable localization:
  {{Anywhere}} and anytime,'' \emph{IEEE Signal Processing Magazine}, vol.~34,
  no.~2, pp. 60--69, 2017.

\bibitem{wuVINS2017}
K.~Wu, C.~Guo, G.~Georgiou \emph{et~al.}, ``{{VINS}} on {{Wheels}},'' in
  \emph{International {{Conference}} on {{Robotics}} and {{Automation}}}.\hskip
  1em plus 0.5em minus 0.4em\relax {IEEE}, 2017, pp. 5155--5162.

\bibitem{brunkerOdometry2019}
A.~Brunker, T.~Wohlgemuth, M.~Frey \emph{et~al.}, ``Odometry 2.0: {{A
  Slip}}-{{Adaptive EIF}}-{{Based Four}}-{{Wheel}}-{{Odometry Model}} for
  {{Parking}},'' \emph{IEEE Transactions on Intelligent Vehicles}, vol.~4,
  no.~1, pp. 114--126, 2019.

\bibitem{zhengSE2018}
F.~Zheng and Y.-H. Liu, ``\BIBforeignlanguage{en}{{{SE}}(2)-{{Constrained
  Visual Inertial Fusion}} for {{Ground Vehicles}}},''
  \emph{\BIBforeignlanguage{en}{IEEE Sensors Journal}}, vol.~18, no.~23, pp.
  9699--9707, 2018.

\bibitem{buczkoSelfValidation2018}
M.~Buczko, V.~Willert, J.~Schwehr \emph{et~al.}, ``Self-{{Validation}} for
  {{Automotive Visual Odometry}},'' in \emph{Intelligent {{Vehicles
  Symposium}}}.\hskip 1em plus 0.5em minus 0.4em\relax {IEEE}, 2018, pp. 1--6.

\bibitem{kokUsing2017}
M.~Kok, J.~D. Hol, and T.~B. Sch\"on, ``Using {{Inertial Sensors}} for
  {{Position}} and {{Orientation Estimation}},'' \emph{Foundations and
  Trends\textregistered{} in Signal Processing}, vol.~11, no. 1-2, pp. 1--153,
  2017.

\bibitem{ramanandanInertial2012}
A.~Ramanandan, A.~Chen, and J.~Farrell, ``Inertial {{Navigation Aiding}} by
  {{Stationary Updates}},'' \emph{IEEE Transactions on Intelligent
  Transportation Systems}, vol.~13, no.~1, pp. 235--248, 2012.

\bibitem{dissanayakeAiding2001}
G.~Dissanayake, S.~Sukkarieh, E.~Nebot \emph{et~al.}, ``The {{Aiding}} of a
  {{Low}}-cost {{Strapdown Inertial Measurement Unit Using Vehicle Model
  Constraints}} for {{Land Vehicle Applications}},'' \emph{IEEE Transactions on
  Robotics and Automation}, vol.~17, no.~5, pp. 731--747, 2001.

\bibitem{brossardRINSW2019}
\BIBentryALTinterwordspacing
M.~Brossard, A.~Barrau, and S.~Bonnabel,
  ``\BIBforeignlanguage{en}{{{RINS}}-{{W}}: {{Robust Inertial Navigation
  System}} on {{Wheels}}},'' \emph{\BIBforeignlanguage{en}{Submitted to IROS
  2019}}, 2019. [Online]. Available:
  \url{https://hal.archives-ouvertes.fr/hal-02057117/file/main.pdf}
\BIBentrySTDinterwordspacing

\bibitem{aghiliRobust2016}
F.~Aghili and C.-Y. Su, ``\BIBforeignlanguage{en}{Robust {{Relative
  Navigation}} by {{Integration}} of {{ICP}} and {{Adaptive Kalman Filter Using
  Laser Scanner}} and {{IMU}}},'' \emph{\BIBforeignlanguage{en}{IEEE/ASME
  Transactions on Mechatronics}}, vol.~21, no.~4, pp. 2015--2026, 2016.

\bibitem{yanRIDI2018}
H.~Yan, Q.~Shan, and Y.~Furukawa, ``{{RIDI}}: {{Robust IMU Double
  Integration}},'' in \emph{European {{Conference}} on {{Computer Vision}}},
  2018.

\bibitem{chenIONet2018}
C.~Chen, X.~Lu, A.~Markham \emph{et~al.}, ``{{IONet}}: {{Learning}} to {{Cure}}
  the {{Curse}} of {{Drift}} in {{Inertial Odometry}},'' in \emph{Conference on
  {{Artificial Intelligence AAAI}}}, 2018.

\bibitem{haarnojaBackprop2016}
T.~Haarnoja, A.~Ajay, S.~Levine \emph{et~al.},
  ``\BIBforeignlanguage{English}{Backprop {{KF}}: {{Learning Discriminative
  Deterministic State Estimators}}},'' in
  \emph{\BIBforeignlanguage{English}{Advances in {{Neural Information
  Processing Systems}}}}, 2016.

\bibitem{liuDeep2018}
K.~Liu, K.~Ok, W.~{Vega-Brown} \emph{et~al.}, ``\BIBforeignlanguage{en}{Deep
  {{Inference}} for {{Covariance Estimation}}: {{Learning Gaussian Noise
  Models}} for {{State Estimation}}},'' in
  \emph{\BIBforeignlanguage{en}{International {{Conference}} on {{Robotics}}
  and {{Automation}}}}.\hskip 1em plus 0.5em minus 0.4em\relax {IEEE}, 2018.

\bibitem{brossardLearning2019}
M.~Brossard and S.~Bonnabel, ``\BIBforeignlanguage{en}{Learning {{Wheel
  Odometry}} and {{IMU Errors}} for {{Localization}}},'' in
  \emph{\BIBforeignlanguage{en}{International {{Conference}} on {{Robotics}}
  and {{Automation}}}}.\hskip 1em plus 0.5em minus 0.4em\relax {IEEE}, 2019.

\bibitem{castellaAdaptive1980}
F.~Castella, ``\BIBforeignlanguage{en}{An {{Adaptive Two}}-{{Dimensional Kalman
  Tracking Filter}}},'' \emph{\BIBforeignlanguage{en}{IEEE Transactions on
  Aerospace and Electronic Systems}}, vol. AES-16, no.~6, pp. 822--829, 1980.

\bibitem{abbeelDiscriminative2005}
P.~Abbeel, A.~Coates, M.~Montemerlo \emph{et~al.}, ``Discriminative
  {{Training}} of {{Kalman Filters}},'' in \emph{Robotics: {{Science}} and
  Systems}, vol.~2, 2005.

\bibitem{tahkTarget1990}
M.~Tahk and J.~Speyer, ``Target tracking problems subject to kine-matic
  constraints,'' \emph{IEEE Transactions on Automatic Control}, vol.~32, no.~2,
  pp. 324--326, 1990.

\bibitem{simonKalman2010}
D.~Simon, ``Kalman {{Filtering}} with {{State Constraints}}: A {{Survey}} of
  {{Linear}} and {{Nonlinear Algorithms}},'' \emph{IET Control Theory \&
  Applications}, vol.~4, no.~8, pp. 1303--1318, 2010.

\bibitem{barrauAligment2016}
A.~Barrau and S.~Bonnabel, ``Aligment {{Method}} for an {{Inertial Unit}},''
  Patent 15/037,653, 2016.

\bibitem{brossardUnscented2018}
M.~Brossard, S.~Bonnabel, and A.~Barrau, ``Unscented {{Kalman Filter}} on {{Lie
  Groups}} for {{Visual Inertial Odometry}},'' in \emph{International
  {{Conference}} on {{Intelligent Robots}} and {{Systems}}}.\hskip 1em plus
  0.5em minus 0.4em\relax {IEEE/RSJ}, 2018.

\bibitem{heoConsistent2018}
S.~Heo and C.~G. Park, ``Consistent {{EKF}}-{{Based Visual}}-{{Inertial
  Odometry}} on {{Matrix Lie Group}},'' \emph{IEEE Sensors Journal}, vol.~18,
  no.~9, pp. 3780--3788, 2018.

\bibitem{hartleyContactAided2018}
R.~Hartley, M.~G. Jadidi, J.~W. Grizzle \emph{et~al.}, ``Contact-{{Aided
  Invariant Extended Kalman Filtering}} for {{Legged Robot State
  Estimation}},'' in \emph{Robotics {{Science}} and {{Systems}}}, 2018.

\bibitem{goodfellowDeep2016}
I.~Goodfellow, Y.~Bengio, and A.~Courville, \emph{Deep {{Learning}}}.\hskip 1em
  plus 0.5em minus 0.4em\relax {The MIT press}, 2016.

\bibitem{kingmaAdam2014}
D.~P. Kingma and J.~Ba, ``{{ADAM}}: {{A Method}} for {{Stochastic
  Optimization}},'' in \emph{International {{Conference}} on {{Learning
  Representations}}}, 2014.

\bibitem{parisiContinual2019}
G.~Parisi, R.~Kemker, J.~Part \emph{et~al.},
  ``\BIBforeignlanguage{en}{Continual {{Lifelong Learning}} with {{Neural
  Networks}}: A {{Review}}},'' \emph{\BIBforeignlanguage{en}{Neural Networks}},
  vol. 113, pp. 54--71, 2019.

\bibitem{qinGeneral2019}
T.~Qin, J.~Pan, S.~Cao \emph{et~al.}, ``\BIBforeignlanguage{en}{A {{General
  Optimization}}-based {{Framework}} for {{Local Odometry Estimation}} with
  {{Multiple Sensors}}},'' 2019.

\bibitem{ramezaniVehicle2018}
M.~Ramezani and K.~Khoshelham, ``\BIBforeignlanguage{en}{Vehicle
  {{Positioning}} in {{GNSS}}-{{Deprived Urban Areas}} by {{Stereo
  Visual}}-{{Inertial Odometry}}},'' \emph{\BIBforeignlanguage{en}{IEEE
  Transactions on Intelligent Vehicles}}, vol.~3, no.~2, pp. 208--217, 2018.

\bibitem{heoEKFBased2018}
S.~Heo, J.~Cha, and C.~G. Park, ``{{EKF}}-{{Based Visual Inertial Navigation
  Using Sliding Window Nonlinear Optimization}},'' \emph{IEEE Transactions on
  Intelligent Transportation Systems}, pp. 1--10, 2018.

\bibitem{sunderhauflimits2018}
N.~S\"underhauf, O.~Brock, W.~Scheirer \emph{et~al.},
  ``\BIBforeignlanguage{en}{The limits and potentials of deep learning for
  robotics},'' \emph{\BIBforeignlanguage{en}{The International Journal of
  Robotics Research}}, vol.~37, no. 4-5, pp. 405--420, 2018.

\end{thebibliography}

	\appendices

	\section{}\label{sec:iekf_prop2}
	The Invariant Extended Kalman Filter (IEKF) \cite{barrauInvariant2017,barrauInvariant2018} is an EKF based on an alternative  state error. One must define a linearized error, an underlying group to derive the exponential map, and then   methodology is akin to the EKF's.

	\subsubsection{Linearized Error}
	the filter state $\bfx_n$ is given by \eqref{statedef}. The state evolution is given by  the dynamics \eqref{eq:prop1}-\eqref{eq:prop3} and \eqref{eq:ass3}-\eqref{eq:ass4}, see Section \ref{sec:model}.  Along the lines of \cite{barrauInvariant2017}, variables $\bochi_n^{\textsc{imu}}:=\left(\bfR_n^{\textsc{imu}}, \bfv_n^{\textsc{imu}}, \bfp_n^{\textsc{imu}}\right)$ are embedded in the Lie group $SE_2(3)$ 
	(see Appendix \ref{sec:se2_3} for the definition of $SE_2(3)$ and its exponential map). Then  biases vector $\bfb_n = \left[\bfb_{n}^{\boomega T}, \bfb_{n}^{\bfa T} \right]^T \in \bbR^6$ is merely treated as a vector, that is, as an element of $\bbR^6$ viewed as a Lie group endowed with standard addition, $\bfR_n^{\mathsf{c}} $ is treated as element of Lie group $SO(3)$, and $\bfp_n^{\mathsf{c}} \in \mathbb R^3$ as a vector. Once the state is broken into several Lie groups, the linearized error writes as the concatenation of corresponding linearized errors, that is, 
	\begin{align}
	\bfe_n = \begin{bmatrix}\boxi_n^{\textsc{imu}T} & \bfe_n^{\bfb T}& \boxi_n^{\bfR^\mathsf{c}T}& \bfe_n^{\bfp^\mathsf{c}T} \end{bmatrix}^T \sim \calN\left(\bfzero, \bfP_n\right),\label{lineq}
	\end{align}
	where state uncertainty $\bfe_n\in \bbR^{21}$ is a zero-mean Gaussian variable with covariance $\bfP_n \in \bbR^{21\times 21}$. As \eqref{eq:ass2} are measurements expressed in the robot's frame, they lend themselves to the Right IEKF methodology. This means each linearized error is mapped to the state using the corresponding Lie group exponential map, and multiplying it  \emph{on the right} by elements of the state space. This yields:
	\begin{align}
	\bochi_n^{\textsc{imu}} &= \exp_{SE_2(3)}\left(\boxi_n^{\textsc{imu}}\right) \hbochi_n^{\textsc{imu}}, \\
	\bfb_n &= \hbfb_n + \bfe_n^{\bfb}, \\
	\bfR_n^{\mathsf{c}} &= \exp_{SO(3)}\left(\boxi_n^{\bfR^\mathsf{c}}\right) \hbfR_n^{\mathsf{c}}, \\
	\bfp_n^{\mathsf{c}} &= \hbfp_n^{\mathsf{c}} + \bfe_n^{\bfp^\mathsf{c}},
	\end{align}
	where $\hat{(\cdot)}$ denotes estimated state variables.

	\subsubsection{Propagation Step}
	we apply \eqref{eq:prop1}-\eqref{eq:prop3} and \eqref{eq:ass3}-\eqref{eq:ass4} to propagate the state and obtain $\hbfx_{n+1}$ and associated covariance through the Riccati equation  
	\eqref{eq:covprop}    
	where the Jacobians $\bfF_n$, $\bfG_n$ are related to the evolution of error \eqref{lineq} and write:
	\begin{align}
	\begin{split}
	\bfF_n &= \bfI_{21\times 21} + \\ &\begin{bmatrix}
	\bfzero & \bfzero & \bfzero & -\bfR_n & \bfzero & \bfzero_{3\times 6}  \\
	(\bfg)_\times & \bfzero & \bfzero & -(\bfv_n^{\textsc{imu}})_\times\bfR_n & -\bfR_n  & \bfzero_{3\times 6}  \\
	\bfzero & \bfI_3 & \bfzero & -(\bfp_n^{\textsc{imu}})_\times \bfR_n & \bfzero & \bfzero_{3\times 6}  \\
	&  \multicolumn{3}{c}{\bfzero_{12 \times 21}} &  
	\end{bmatrix} dt, 
	\end{split}\label{eq:F}
	\end{align}
	\begin{align}
	\bfG_n = \begin{bmatrix}
	\bfR_n & \bfzero & \bfzero_{3\times 12}  \\
	(\bfv_n^{\textsc{imu}})_\times \bfR_n & \bfR_n & \bfzero_{3\times 12}   \\
	(\bfp_n^{\textsc{imu}})_\times \bfR_n & \bfzero & \bfzero_{3\times 12}  \\
	\bfzero_{12\times 3} & \bfzero_{12\times 3} & \bfI_{12 \times 12}
	\end{bmatrix} dt,
	\end{align}
	with $\bfR_n=\bfR_n^{\textsc{imu}}$, $\bfzero=\bfzero_{3\times 3}$, and   $\bfQ_n$ denotes the classical covariance matrix of the process noise as in Section \ref{sec:implementation}. 
	\subsubsection{Update Step} 
	the measurement vector $\bfy_{n+1}$ is computed by stacking the motion information
	\begin{align}
	\bfy_{n+1} &= \begin{bmatrix}
	v_{n+1}^{\mathrm{lat}} \\
	v_{n+1}^{\mathrm{up}}
	\end{bmatrix} = \bfzero, \label{eq:y_lat}
	\end{align}
	with assessed uncertainty 
	a zero-mean Gaussian variable with covariance $\bfN_{n+1} = \cov\left(\bfy_{n+1}\right)$. We then compute an updated state $\hbfx_{n+1}^+$ and updated covariance $\bfP_{n+1}^+$ following the IEKF methodology, i.e. we compute
	\begin{align}
	\bfS &= \left(\bfH_{n+1} \bfP_{n+1} \bfH_{n+1}^T + \bfN_{n+1}\right), \\
	\bfK &= \bfP_{n+1} \bfH_{n+1}^T/\bfS, \label{eq:gain}\\
	\bfe^+ &= \bfK\left(\bfy_{n+1}-\hbfy_{n+1}\right) \label{eq:innovation}, \\
	\hbochi_{n+1}^{\textsc{imu}+} &= \exp_{SE_2(3)}\left(\boxi^{\textsc{imu}+}\right) \hbochi_{n+1}^{\textsc{imu}},\label{eq:upstate1} \\
	\bfb_{n+1}^+ &= \bfb_{n+1} + \bfe^{\bfb+} \\
	\hbfR_{n+1}^{\mathsf{c}+} &= \exp_{SO(3)}\left(\boxi^{\bfR^\mathsf{c}+}\right) \hbfR_{n+1}^{\mathsf{c}}, \\ \hbfp_{n+1}^{\mathsf{c}+} &= \hbfp_{n+1}^{\mathsf{c}} + \bfe^{\bfp^\mathsf{c}+}, 
	\label{eq:upstate2}\\
	\bfP_{n+1}^+ &= \left(\bfI_{21} - \bfK \bfH_{n+1}\right) \bfP_{n+1}, \label{eq:upcov}
	\end{align}
	summarized as Kalman gain \eqref{eq:gain}, state innovation \eqref{eq:innovation}, state update \eqref{eq:upstate1}-\eqref{eq:upstate2} and covariance update \eqref{eq:upcov}, where  $\bfH_{n+1}$ is  the measurement Jacobian matrix with respect to linearized error \eqref{lineq} and thus given as:
	\begin{align}
	\bfH_n &= \bfA \begin{bmatrix}
	\bfzero & \bfR_n^{\textsc{imu}T} & \bfzero & -\left(\bfp_n^{\mathsf{c}}\right)_\times & \bfzero & \bfB & \bfC
	\end{bmatrix}\label{eq:H},
	\end{align}
	where $\bfA = \left[\bfI_2~ \bfzero_{2}\right]$ selects the two first row of the right part of \eqref{eq:H}, $\bfB=\bfR_n^{\mathsf{c}T} \bfR_n^{\textsc{imu}T} \left(\bfv_n^{\textsc{imu}}\right)_\times$ and $\bfC =-(\boomega_n^{\textsc{imu}}-\bfb_n^{\boomega})_\times$.
	
	\section{}
	\label{sec:se2_3}
	The Lie group $SE_2(3)$ is an extension of the Lie group $SE(3) $ and is described as follows, see  \cite{barrauInvariant2017} where it was first introduced. A $5\times5$ matrix $\bochi_n \in SE_{2}(3)$ is defined as
	\begin{align}
	\bochi_n =  \left[\begin{array}{ccc}
	\bfR_n & \bfv_n & \bfp_n \\
	\bfzero_{2\times 3} & \multicolumn{2}{c}{\bfI_2}
	\end{array} \right] \in SE_2(3).\label{bochi:eq}
	\end{align}
	The uncertainties $\boxi_n \in \bbR^{9}$ are mapped to the Lie algebra $\mathfrak{se}_2(3)$ through the transformation $\boxi_n\mapsto\boxi_n^{\wedge} $ defined as
	\begin{align}
	\boxi_n &= \left[\boxi^{\bfR T}_n,~ \boxi^{\bfv T}_n,~ \boxi^{\bfp T}_n \right]^T, \\
	\boxi^{\wedge}_n &= \left[\begin{array}{ccc} \left(\boxi^\bfR_n\right)_\times & \boxi^\bfv_n & \boxi^\bfp_n  \\
	\multicolumn{3}{c}{\bfzero_{2\times 5}}
	\end{array} \right] \in \mathfrak{se}_2(3),
	\end{align}
	where $\boxi^\bfR_n \in \bbR^3$, $\boxi^\bfv_n \in \bbR^3$ and $\boxi^\bfp_n \in \bbR^3$. The closed-form expression for the exponential map is given as
	\begin{align}
	\exp_{SE_2(3)}\left(\boxi_n\right) = \bfI + \boxi_n^\wedge + a (\boxi_n^{\wedge})^{2} + b (\boxi_n^{\wedge})^{3},
	\end{align}
	where $a=\frac{1-\cos\left(\|\boxi^\bfR_n\|\right)}{\|\boxi^\bfR_n\|}$ and $b=\frac{ \|\boxi^\bfR_n\|-\sin\left(\|\boxi^\bfR_n\|\right)}{\|\boxi^\bfR_n\|^3}$, and which inherently uses the exponential of $SO(3)$, defined as
	\begin{align}
	\exp_{SO(3)}\left(\boxi^\bfR_n\right) &= \exp\left(\left(\boxi^\bfR_n\right)_\times\right) \\
	&= \bfI + \left(\boxi^\bfR_n\right)_\times + a \left(\boxi^\bfR_n\right)_\times^{2}. \label{eq:so3}
	\end{align}
	
\end{document}